\documentclass{article} 
\PassOptionsToPackage{dvipsnames}{xcolor}
\usepackage{iclr2019_conference,times}

\usepackage{amsmath,amsfonts,bm}

\def\eqref#1{equation~\ref{#1}}

\def\1{\bm{1}}

\DeclareMathAlphabet{\mathsfit}{\encodingdefault}{\sfdefault}{m}{sl}
\SetMathAlphabet{\mathsfit}{bold}{\encodingdefault}{\sfdefault}{bx}{n}

\usepackage{url}
\usepackage[dvipsnames]{xcolor}
\usepackage{mathtools}
\usepackage{amsmath}
\usepackage{amsthm}
\usepackage{enumerate}
\usepackage{graphicx}
\usepackage[export]{adjustbox}
\usepackage{booktabs}
\usepackage{tikz}
\usepackage{pgfplots}
\usepackage{listings}
\usepackage{subcaption}
\usepackage[normalem]{ulem}
\usepackage{doi}
\useunder{\uline}{\ul}{}
\usepackage[toc,page]{appendix}
\usepackage{paralist}
\usepackage{cleveref}
\usepackage{scalerel}
\usepackage{multirow}
\usepackage[frozencache]{minted}
\usepackage{makecell}
\usepackage{hyperref}
\usepackage{enumitem}

\newcommand{\sname}[1]{{\small \textsc{#1}}}
\newcommand{\scode}[1]{{\small \texttt{#1}}}
\renewcommand\cite{\citep}

\newcommand\para[1]{\noindent \textbf{#1}}

\usepackage[compact]{titlesec}

\title{code2seq: Generating Sequences from \\Structured Representations of Code}

\author{Uri Alon \\
Technion\\
\texttt{urialon@cs.technion.ac.il} \\
\And
Shaked Brody \\
Technion \\
\texttt{shakedbr@cs.technion.ac.il} \\
\And
Omer Levy \\
Facebook AI Research \;\;\;\;\;\;\;\;\;\;\;\;\;\;\;\;\;\;\;\;\;\;\;\;\;\;\;\;\;\;\;\;\;\;\;\;\;\;\;\;\\
\texttt{omerlevy@gmail.com}  \\
\And
Eran Yahav \\
Technion \\
\texttt{yahave@cs.technion.ac.il}
}

\iclrfinalcopy 
\begin{document}

\maketitle

\begin{abstract}
The ability to generate natural language sequences from source code snippets has a variety of applications such as code summarization, documentation, and retrieval. Sequence-to-sequence (seq2seq) models, adopted from neural machine translation (NMT), have achieved state-of-the-art performance on these tasks by treating source code as a sequence of tokens. We present \sname{code2seq}: an alternative approach that leverages the syntactic structure of programming languages to better encode source code.
Our model represents a code snippet as the set of compositional paths in its abstract syntax tree (AST) and uses attention to select the relevant paths while decoding. 
We demonstrate the effectiveness of our approach for two tasks, two programming languages, and four datasets of up to $16$M examples. Our model significantly outperforms previous models that were specifically designed for programming languages, as well as state-of-the-art NMT models. An online demo of our model is available at \url{http://code2seq.org}. Our code, data and trained models are available at \url{http://github.com/tech-srl/code2seq}.
\end{abstract}

\section{Introduction}\label{Intro}







Modeling the relation between source code and natural language can be used for automatic code summarization \cite{conv16}, documentation \cite{codenn16}, retrieval \cite{bimodal15}, and even generation \cite{balog2016deepcoder, rabinovich2017, yin2017, devlin2017robustfill, murali2018bayou,brockschmidt2018generative}. In this work, we consider the general problem of generating a natural language sequence from a given snippet of source code.

A direct approach is to frame the problem as a machine translation problem, where the source sentence is the sequence of tokens in the code and the target sentence is a corresponding natural language sequence.
This approach allows one to apply state-of-the-art neural machine translation (NMT) models from the sequence-to-sequence (seq2seq) paradigm \cite{sutskever2014sequence, cho2014learning, bahdanau14, luong15, vaswani2017attention}, yielding state-of-the-art performance on various code captioning and documentation benchmarks \cite{codenn16, conv16, loyola2017} despite having extremely long source sequences.

We present an alternative approach for encoding source code that leverages the syntactic structure of programming languages: \sname{code2seq}. We represent a given code snippet as a set of compositional paths over its abstract syntax tree (AST), where each path is compressed to a fixed-length vector using LSTMs \cite{hochreiter1997lstm}. During decoding, \sname{code2seq} attends over a different weighted average of the path-vectors to produce each output token, much like NMT models attend over token representations in the source sentence. 

\begin{figure*}[t]

\centering
\begin{minipage}{\textwidth}
\hspace{-3mm}
\begin{tabular}{ll}
\begin{subfigure}[t]{0.41\textwidth}
\centering
\textbf{Code summarization in Java:}
\end{subfigure}

\begin{subfigure}[t]{0.50\textwidth}
\centering
\hspace{0.1cm}
\textbf{Code captioning in C\#:}
\end{subfigure}
\\
\begin{subfigure}[t]{0.41\textwidth}
\frame{\includegraphics[height=2.9cm,keepaspectratio]{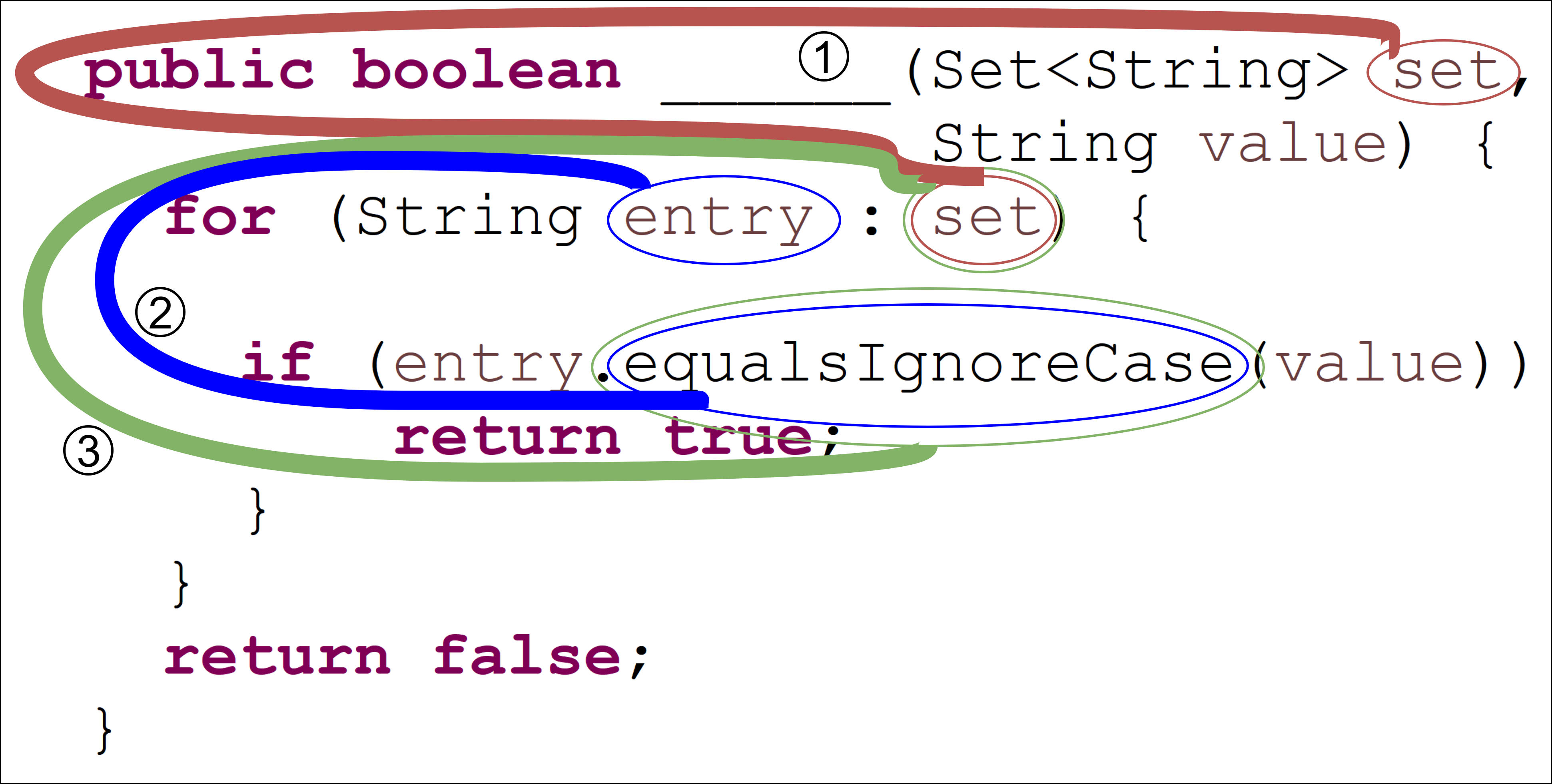}}

\end{subfigure}

\begin{subfigure}[t]{0.50\textwidth}
\frame{\includegraphics[height=2.9cm,keepaspectratio]{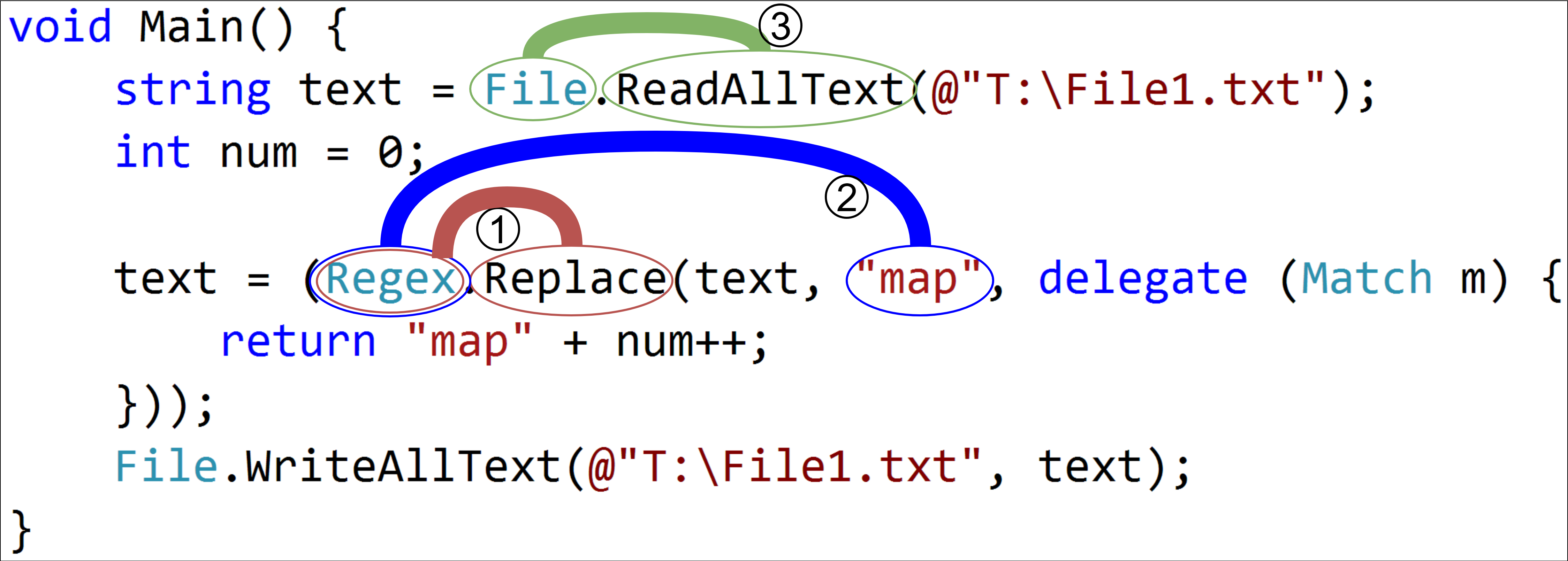}}
\end{subfigure}
\\

\begin{subfigure}[t]{0.41\textwidth}
\centering
\includegraphics[height=0.6cm,keepaspectratio]{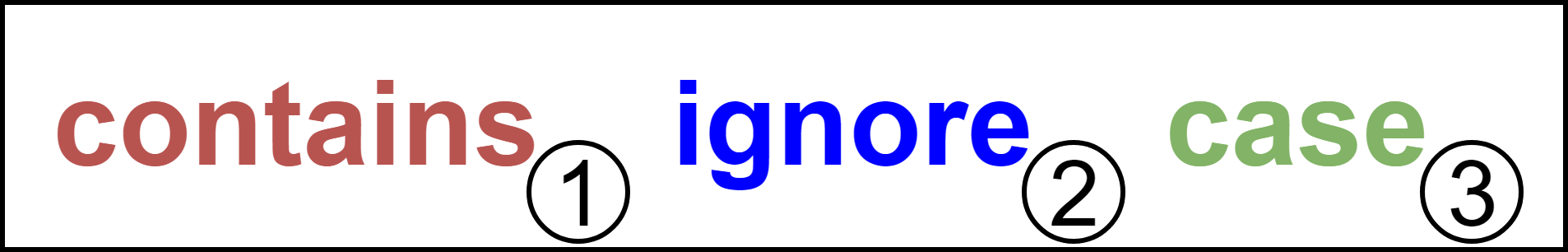}
\caption{}
\label{java_example}
\end{subfigure}

\begin{subfigure}[t]{0.50\textwidth}
\centering
\hspace{10mm}
\includegraphics[height=0.6cm,keepaspectratio]{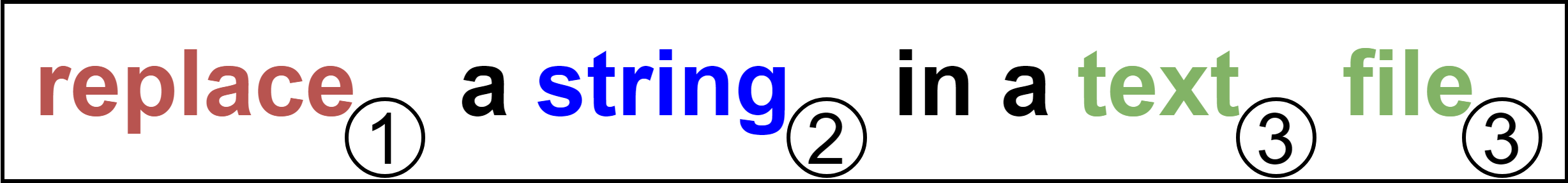}
\caption{}
\label{csharp_example}
\end{subfigure}
\end{tabular}

\end{minipage}
\caption{Example of (a) code summarization of a Java code snippet, and (b) code captioning of a C\# code snippet, along with the predictions produced by our models. The highlighted paths in each example are the top-attended paths in each decoding step. Because of space limitations we included only the top-attended path for each decoding step, but hundreds of paths are attended at each step.
Additional examples are presented in \Cref{appendix_csharp} and \Cref{appendix_java}.}
\label{fig:overview-all}
\end{figure*}

We show the effectiveness of our code2seq model on two tasks: (1) code summarization (\Cref{java_example}), where we predict a Java method's name given its body, and (2) code captioning (\Cref{csharp_example}), where we predict a natural language sentence that describes a given C\# snippet.

On both tasks, our \sname{code2seq} model outperforms models that were explicitly designed for code, such as the model of \citet{conv16}
and CodeNN \cite{codenn16}, as well as TreeLSTMs \cite{tai2015improved} and state-of-the-art NMT models \cite{luong15,vaswani2017attention}.
To examine the importance of each component of the model, we conduct a thorough ablation study. In particular, we show the importance of structural encoding of code, by showing how our model yields a significant improvement over an ablation that uses only token-level information without syntactic paths. To the best of our knowledge, this is the first work to directly use paths in the abstract syntax tree for end-to-end generation of sequences.

\section{Representing Code as AST Paths}\label{paths}

\begin{figure*}[t]
\centering
\begin{minipage}{\textwidth}
\hspace{-2mm}
\begin{tabular}{ll}
\begin{subfigure}[t]{0.46\textwidth}

\begin{minted}[fontsize=\scriptsize, frame=single,framesep=2pt]{Java}
int countOccurrences(String str, char ch) {
   int num = 0;
   int index = -1;
   do {
      index = str.indexOf(ch, index + 1);
      if (index >= 0) {
         num++;
      }
   } while (index >= 0);
   return num;
}
\end{minted}
\caption{}
\label{countOccurrences-a}
\end{subfigure}

\begin{subfigure}[t]{0.53\textwidth}

\begin{minted}[fontsize=\scriptsize,stripnl=false, frame=single,framesep=2pt]{Java}
int countOccurrences(String source, char value) {
   int count = 0;
   for (int i = 0; i < source.length(); i++) {
       if (source.charAt(i) == value) {
           count++;
        }
   }
   return count;
}


\end{minted}

\caption{}
\label{countOccurrences-b}

\end{subfigure}

\\
\\
\begin{subfigure}[b]{0.46\textwidth}
\centering
\includegraphics[scale=0.09]{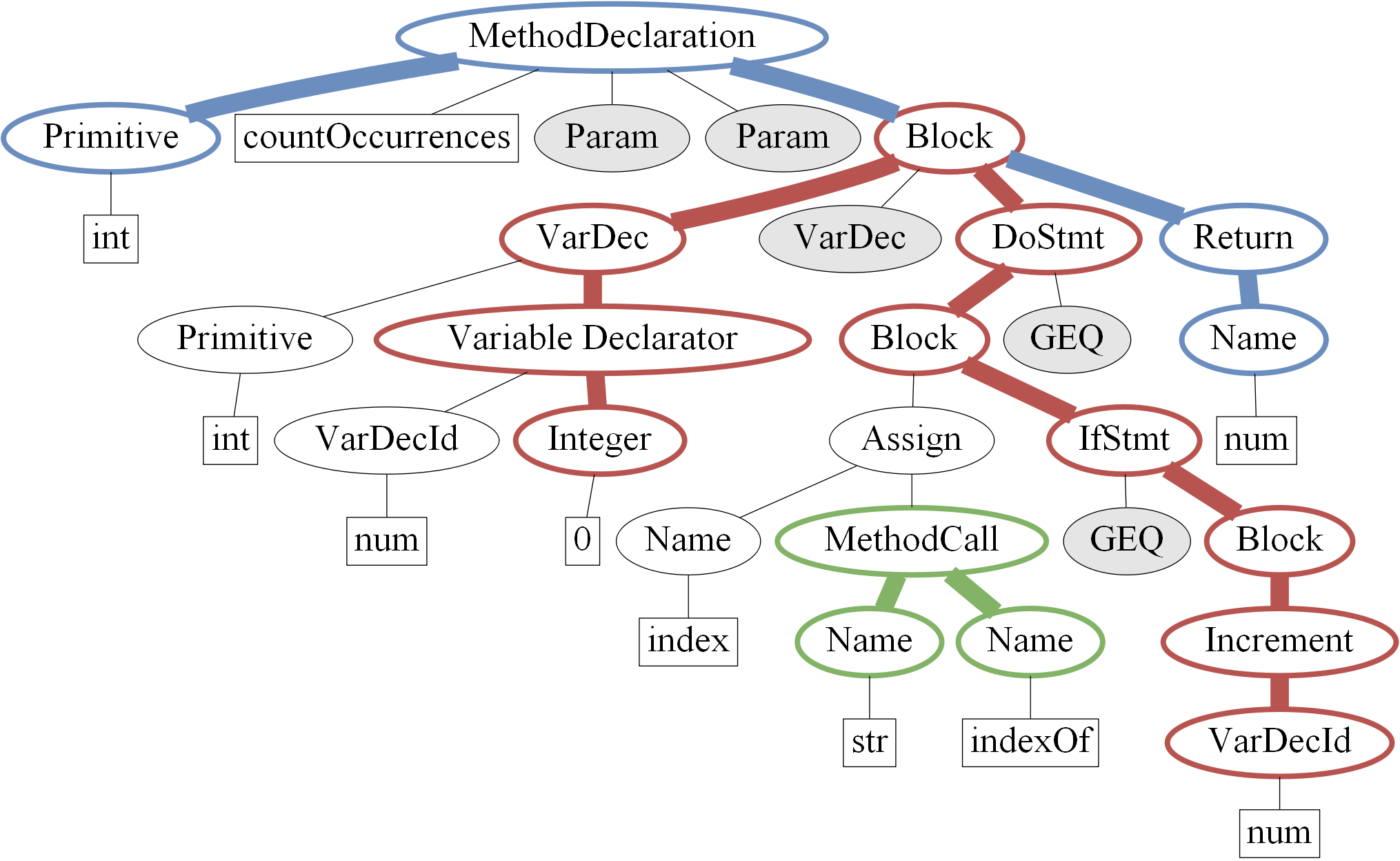}
\caption{}
\label{ast-a}
\end{subfigure}
\begin{subfigure}[b]{0.53\textwidth}
\centering
\includegraphics[scale=0.09]{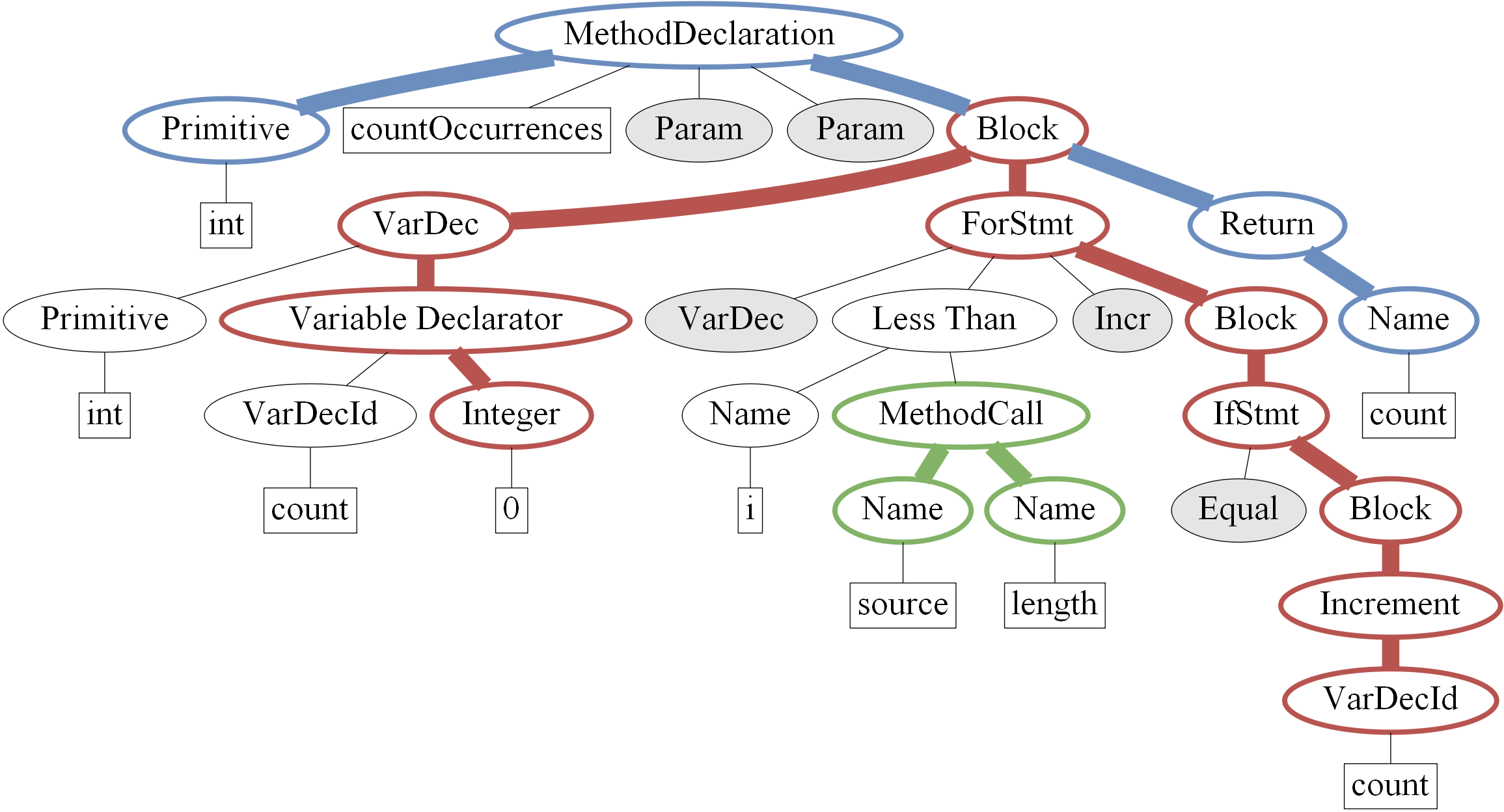}
\caption{}
\label{ast-b}
\end{subfigure}

\end{tabular}
\end{minipage}
\caption{An example of two Java methods that have exactly the same functionality. Although these methods have different \emph{sequential} (token-based) representations, repeating paths, which might differ in only a single node
(a \scode{ForStmt} node instead of a \scode{Do}-while node), will be revealed if we consider syntactic patterns.
}
\label{countOccurrences-all}

\vspace{-5mm}
\end{figure*}

An Abstract Syntax Tree (AST) uniquely represents a source code snippet in a given language and grammar. The leaves of the tree are called \emph{terminals}, and usually refer to user-defined values which represent identifiers and names from the code. The non-leaf nodes are called \emph{nonterminals} and represent a restricted set of structures in the language, e.g., loops, expressions, and variable declarations.
For example, \Cref{ast-a} shows a partial AST for the code snippet of \Cref{countOccurrences-a}.   Names (such as \scode{num}) and types (such as \scode{int}) are represented as values of terminals; syntactic structures such as variable declaration (\scode{VarDec}) and a do-while loop (\scode{DoStmt}) are represented as nonterminals.

Given the AST of a code snippet, we consider all pairwise paths between terminals, and represent them as sequences of terminal and nonterminal nodes. We then use these paths with their terminals' values to represent the code snippet itself.
For example, consider the two Java methods of \Cref{countOccurrences-all}. Both of these methods count occurrences of a character in a string. They have exactly the same functionality, although a different implementation, and therefore different surface forms. If these snippets are encoded as sequences of tokens, the recurring patterns that suggest the common method name might be overlooked. However, a structural observation reveals syntactic paths that are common to both methods, and differ only in a single node of a \scode{Do}-while statement versus a \scode{For} statement. This example shows the effectiveness of a syntactic encoding of code. Such an encoder can generalize much better to unseen examples because the AST normalizes a lot of the surface form variance. Since our encoding is compositional, the encoder can generalize even if the paths are not identical (e.g., a \scode{For} node in one path and a \scode{While} in the other).

Since a code snippet can contain an arbitrary number of such paths, we sample $k$ paths as the representation of the code snippet. To avoid bias, $k$ new paths are sampled afresh in every training iteration. In \Cref{sec:analysis} we show that this runtime-sampling provides regularization and improves results compared to sampling the same $k$ paths for each example in advance.

Formally, we use $\mathcal{C}$ to denote a given snippet of code.
Every training iteration, $k$ pairs of terminals are uniformly sampled from within the AST of $\mathcal{C}$. Each pair of terminals $\left(v_{1}^i,v_{l_i}^i\right)$ implies a single path between them: $v_{1}^{i}v_{2}^{i}...v_{l_i}^{i}$. Finally, the input code example is represented as a set of these $k$ random AST paths: $\left\{\left(v_{1}^{1}v_{2}^{1}...v_{l_1}^{1}\right),...,\left(v_{1}^{k}v_{2}^{k}...v_{l_k}^{k}\right)\right\}$, where $l_j$ is the length of the $j$th path.

\section{Model Architecture}\label{Model}

\begin{figure}
\centering
\includegraphics[width=0.8\textwidth]{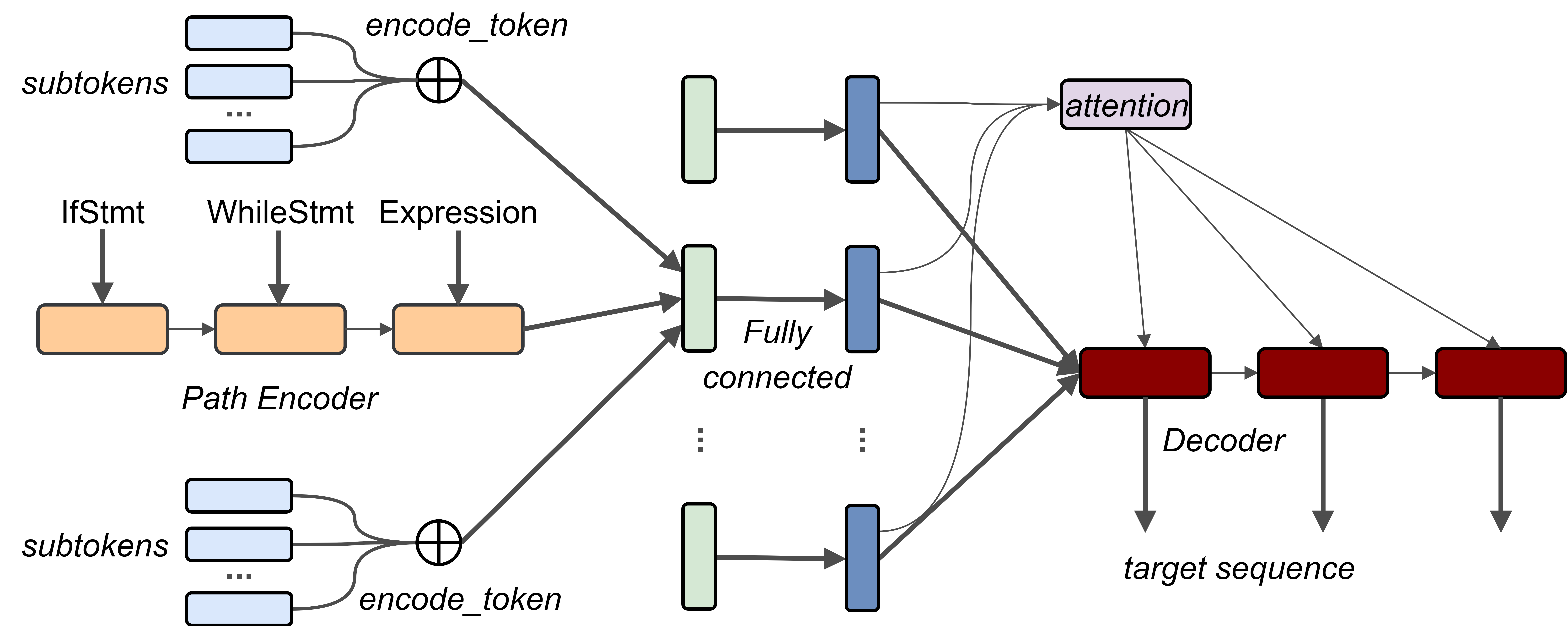}
\caption{Our model encodes each AST path with its values as a vector, and uses the average of all of the $k$ paths as the decoder's start state. The decoder generates an output sequence while attending over the $k$ encoded paths.}
\label{network}
\end{figure}

Our model follows the standard encoder-decoder architecture for NMT (Section~\ref{subsec:nmt}),
with the significant difference that \emph{the encoder does not read the input as a flat sequence of tokens}. Instead, the encoder creates a vector representation for each AST path separately (Section~\ref{subsec:ast_encoder}). The decoder then attends over the encoded AST paths (rather than the encoded tokens) while generating the target sequence. Our model is illustrated in \Cref{network}.

\subsection{Encoder-Decoder Framework}
\label{subsec:nmt}

Contemporary NMT models are largely based on an encoder-decoder architecture \cite{cho2014learning, sutskever2014sequence, luong15, bahdanau14}, where the encoder maps an input sequence of tokens $\boldsymbol{x}=\left(x_1,...,x_n\right)$ to a sequence of continuous representations $\boldsymbol{z}=\left(z_1,...,z_n\right)$. Given $\boldsymbol{z}$, the decoder then generates a sequence of output tokens $\boldsymbol{y}=\left(y_1,...,y_m\right)$ one token at a time, hence modeling the conditional probability:
$p\left(y_1,...,y_m|x_1,...,x_n\right)$.

At each decoding step, the probability of the next target token depends on the previously generated token, and can therefore be factorized as:
\begin{equation*}
p\left(y_1,...,y_m|x_1,...,x_n\right)=\prod_{j=1}^{m}p\left(y_j|y_{<j},z_1,...,z_n\right)
\end{equation*}

In attention-based models, at each time step $t$ in the decoding phase, a context vector $c_{t}$ is computed by attending over the elements in $\boldsymbol{z}$ using the decoding state $h_t$, typically computed by an LSTM.

\begin{equation*}
\boldsymbol{\alpha}^t= softmax \left( h_t W_a \boldsymbol{z} \right)
\qquad
c_t= \sum_i^n \alpha_i^t z_i
\end{equation*}

The context vector  $c_{t}$ and the decoding state $h_{t}$ are then combined to predict the current target token $y_t$. Previous work differs in the way the context vector is computed and in the way it is combined with the current decoding state. A standard approach \cite{luong15} is to pass $c_t$ and $h_t$ through a multi-layer perceptron (MLP) and then predict the probability of the next token using softmax:
\begin{equation*}
p\left(y_t|y_{<t},z_1,...,z_n\right) = softmax\left(W_s tanh\left(W_c\left[c_t;h_t\right]\right) \right)
\end{equation*}

\subsection{AST Encoder}
\label{subsec:ast_encoder}

Given a set of AST paths $\{x_1, ..., x_k\}$, our goal is to create a vector representation $z_i$ for each path $x_i=v_{1}^{i}v_{2}^{i}...v_{l_i}^{i}$. We represent each path separately using a bi-directional LSTM to encode the path, and sub-token embeddings to capture the compositional nature of the terminals' values (the tokens).

\paragraph{Path Representation}
Each AST path is composed of nodes and their child indices from a limited vocabulary of up to $364$ symbols. We represent each node using a learned embedding matrix $E^{nodes}$ and then encode the entire sequence using the final states of a bi-directional LSTM:
\begin{equation*}
h_1,...,h_l = LSTM(E_{v_{1}}^{nodes},...,E_{v_{l}}^{nodes})
\end{equation*}
\begin{equation*}
encode\_path(v_{1}...v_{l}) = \left[ h_l^{\rightarrow} ; h_1^{\leftarrow} \right]
\end{equation*}

\paragraph{Token Representation}
The first and last node of an AST path are terminals whose values are tokens in the code.
Following \citet{allamanis2015, conv16}, we split code tokens into \emph{sub}tokens;
for example, a token with the value \scode{ArrayList} will be decomposed into \scode{Array} and \scode{List}.
This is somewhat analogous to byte-pair encoding in NMT \cite{sennrich2016}, although in the case of programming languages, coding conventions such as camel notation provide us with an explicit partition of each token.
We use a learned embedding matrix $E^{subtokens}$ to represent each subtoken, and then sum the subtoken vectors to represent the full token:
\begin{equation*}
encode\_token(w) = \sum_{s \in split(w)} E_s^{subtokens}
\end{equation*}
The LSTM decoder may also predict subtokens at each step (e.g. when generating method names), although the decoder's subtoken embedding matrix will be different.

\paragraph{Combined Representation}
To represent the path $x=v_{1}...v_{l}$, we concatenate the path's representation with the token representations of each terminal node, and apply a fully-connected layer:
\begin{equation*}
z = tanh \left( W_{in} \left[ encode\_path(v_{1}...v_{l}) ; encode\_token(value(v_{1})) ; encode\_token(value(v_{l})) \right] \right)
\end{equation*}
where $value$ is the mapping of a terminal node to its associated value, and $W_{in}$ is a $(2d_{path} + 2d_{token}) \times d_{hidden}$ matrix.

\paragraph{Decoder Start State}
To provide the decoder with an initial state, we average the combined representations of \emph{all} the $k$ paths in the given example:
\begin{equation*}
h_0 = \frac{1}{k}\sum_{i=1}^{k}z_i
\end{equation*}
Unlike typical encoder-decoder models, the order of the input random paths is not taken into account. Each path is encoded separately and the combined representations are aggregated with mean pooling to initialize the decoder's state. This represents the given source code as a \emph{set} of random paths.

\paragraph{Attention} Finally, the decoder generates the output sequence while \emph{attending over all of the combined representations} $z_1,...z_k$, similarly to the way that seq2seq models attend over the source symbols. The attention mechanism is used to dynamically select the distribution over these $k$ combined representations while decoding, just as an NMT model would attend over the encoded source tokens. 
\section{Experiments}\label{evaluation}

We evaluate our model on two code-to-sequence tasks: summarization (\Cref{subsec:summarization}), in which we predict Java methods' names from their bodies, and captioning (\Cref{subsec:captioning}), where we generate natural language descriptions of C\# code snippets. Although out of the focus of this work, in \Cref{subsec:documentation} we show that our model also generates Javadocs more accurately than an existing work. We thus demonstrate that our approach can produce both method names and natural language outputs, and can encode a code snippet in any language for which an AST can be constructed (i.e., a parser exists).

\para{Setup}
The values of all of the parameters are initialized using the initialization heuristic of \citet{glorot2010understanding}. We optimize the cross-entropy loss~\cite{rubinstein1999cross, rubinstein2001combinatorial} with a Nesterov momentum \cite{nesterov1983method} of $0.95$ and an initial learning rate of $0.01$, decayed by a factor of $0.95$ every epoch. For the Code Summarization task, we apply dropout \cite{srivastava2014dropout} of $0.25$ on the input vectors $x_j$, and $0.7$ for the Code Captioning task because of the smaller number of examples in the C\# dataset. We apply a recurrent dropout of $0.5$ on the LSTM that encodes the AST paths.
We used $d_{tokens}=d_{nodes}=d_{hidden}=d_{target}=128$. For the Code Summarization task, each LSTM that encodes the AST paths had $128$ units and the decoder LSTM had $320$ units. For the Code Captioning task, to support the longer target sequences, each encoder LSTM had $256$ units and the decoder was of size $512$.

\paragraph{Choice of $k$} We experimented with different values of $k$, the number of sampled paths from each example (which we set to $200$ in the final models).
Lower values than $k=100$ showed worse results, and increasing to $k>300$ did not result in consistent improvement.
In practice, $k=200$ was found to be a reasonable sweet spot between capturing enough information while keeping training feasible in the GPU's memory.
Additionally, since the average number of paths in our Java-large training set is $220$ paths per example, a number as high as $200$ is beneficial for some large methods.

\subsection{Code Summarization}
\label{subsec:summarization}

In this task, we predict a Java method's name given its body. As was previously observed \cite{conv16, alon2019code2vec}, this is a good benchmark because a method name in open-source Java projects tends to be succinct and precise, and a method body is often a complete logical unit.
We predict the target method name as a sequence of sub-tokens, e.g., \scode{setMaxConnectionsPerServer} is predicted as the sequence ``set max connections per server''. The target sequence length is about $3$ on average.
We adopt the measure used by \citet{conv16} and \citet{alon2019code2vec}, who measured precision, recall, and F1 score over the target sequence, case insensitive.

\para{Data}
We experiment with this task across three datsets. In these datasets, we always train across multiple projects and predict on distinct projects:

\emph{Java-small} -- Contains $11$ relatively large Java projects, originally used for $11$ distinct models for training and predicting within the scope of the same project \cite{conv16}. We use the same data, but \emph{train and predict across projects}: we took $9$ projects for training, 1 project for validation and $1$ project as our test set. This dataset contains about $700K$ examples.

\emph{Java-med} -- A new dataset of the $1000$ top-starred Java projects from GitHub. We randomly select $800$ projects for training, $100$ for validation and $100$ for testing.
This dataset contains about 4M examples and we make it publicly available.

\emph{Java-large} -- A new dataset of the $9500$ top-starred Java projects from GitHub that were created since January 2007. We randomly select $9000$ projects for training, $250$ for validation and $300$ for testing. This dataset contains about $16M$ examples and we make it publicly available.

More statistics of our datasets can be found in \Cref{appendix_details}.

\para{Baselines}
We re-trained all of the baselines on all of the datasets of this task using the original implementations of the authors.
We compare \sname{code2seq} to the following baselines: \citet{conv16}, who used a convolutional attention network to predict method names; syntactic paths with Conditional Random Fields (CRFs) \cite{pigeon}; code2vec \cite{alon2019code2vec}; and a TreeLSTM \cite{tai2015improved} encoder with an LSTM decoder and attention on the input sub-trees. Additionally, we compared to three NMT baselines that read the input source code as a stream of tokens: $2$-layer bidirectional encoder-decoder LSTMs (split tokens and full tokens) with global attention \cite{luong15}, and the Transformer \cite{vaswani2017attention}, which achieved state-of-the-art results for translation tasks.


We put significant effort into strengthening the NMT baselines in order to provide a fair comparison: (1) we split tokens to subtokens, as in our model (e.g., \scode{HashSet} $\rightarrow$ \scode{Hash Set}) -- this was shown to improve the results by about $10$ F1 points (\Cref{methods-results}); (2) we deliberately kept the original casing of the source tokens since we found it to improve their results; and (3) during inference, we replaced generated UNK tokens with the source tokens that were given the highest attention.
For the 2-layer BiLSTM we used embeddings of size $512$, an encoder and a decoder of $512$ units each, and the default hyperparameters of OpenNMT \cite{2017opennmt}. For the Transformer, we used their original hyperparameters \cite{vaswani2017attention}. This resulted in a Transformer model with $169M$ parameters and a BiLSTM model with $134M$ parameters, while our code2seq model had only 37M.\footnote{We also trained versions of the NMT baselines in which we down-matched the sizes and number of parameters to our model. These baselines seemed to benefit from more parameters, so the results reported here are for the versions that had many more parameters than our model.}




\begin{table}[t]
\caption{Our model significantly outperforms  previous PL-oriented and NMT models. Another visualization can be found in \Cref{java_appendix_section}.}
\label{methods-results}
\scriptsize
\centering
\makebox[\linewidth]{
\begin{tabular}{llllllllll}
\toprule
\multirow{2}{*}{Model}              & \multicolumn{3}{c}{Java-small}    & \multicolumn{3}{c}{Java-med}    & \multicolumn{3}{c}{Java-large}    \\
\cmidrule{2-4} \cmidrule{5-7} \cmidrule{8-10}
                                    & Prec      & Rec         & F1             & Prec       & Rec         & F1             & Prec            & Rec         & F1             \\
\midrule
ConvAttention \cite{conv16}         & 50.25          & 24.62          & 33.05          & 60.82           & 26.75         &  37.16          & 60.71               & 27.60          & 37.95          \\
Paths+CRFs \cite{pigeon}                          & 8.39        & 5.63               & 6.74           & 32.56      & 20.37               & 25.06               & 32.56                     & 20.37               & 25.06          \\
code2vec \cite{alon2019code2vec} & 18.51 &18.74 & 18.62 & 38.12 & 28.31& 32.49& 48.15 & 38.40 & 42.73 \\
2-layer BiLSTM (no token splitting)     &  32.40         & 20.40         & 25.03         & 48.37           & 30.29          & 37.25            & 58.02 & 37.73 & 45.73                               \\
2-layer BiLSTM                             & 42.63          & 29.97          & 35.20          & 55.15           & 41.75          & 47.52          & 63.53         & 48.77               & 55.18 \\
TreeLSTM \cite{tai2015improved} & 40.02 & 31.84 & 35.46 & 53.07 & 41.69 & 46.69 & 60.34 & 48.27 & 53.63 \\
Transformer \cite{vaswani2017attention}                        & 38.13          & 26.70          & 31.41         & 50.11 & 35.01               & 41.22         & 59.13 & 40.58               & 48.13               \\
\midrule 
code2seq & \textbf{50.64}      & \textbf{37.40}      & \textbf{43.02}       & \textbf{61.24} & \textbf{47.07}  & \textbf{53.23}& \textbf{64.03}            & \textbf{55.02}      & \textbf{59.19} \\
Absolute gain over BiLSTM & +8.01      & +7.43      & +7.82       & +6.09 & +5.32  & +5.71 & +0.50            & +6.25      & +4.01 \\
\bottomrule
\end{tabular}
}
\end{table} 

\para{Performance}
\Cref{methods-results} shows the results for the code summarization task.
Our model significantly outperforms the baselines in both precision and recall across all three datasets, demonstrating that there is added value in leveraging ASTs to encode source code. Our model improves over the best baselines, BiLSTM with split tokens, by between $4$ to $8$ F1 points on all benchmarks. BiLSTM with split tokens consistently scored about $10$ F1 points more than BiLSTM with full tokens, and for this reason we included only the split token Transformer and TreeLSTM baselines.
Our model outperforms ConvAttention \cite{conv16}, which was designed specifically for this task; Paths+CRFs \cite{pigeon}, which used syntactic features; and TreeLSTMs. Although TreeLSTMs also leverage syntax, we hypothesize that our syntactic paths capture long distance relationships while TreeLSTMs capture mostly local properties.
An additional comparison to code2vec on the code2vec dataset can be found in \Cref{appendix_details}.
Examples for predictions made by our model and each of the baselines can be found in \Cref{appendix_java} and at \url{http://code2seq.org}.

\citet{fernandes2018structured} encoded code using Graph Neural Networks (GNN), and reported lower performance than our model on Java-large without specifying the exact F1 score. They report slightly higher results than us on Java-small only by extending their GNN encoder with a subtoken-LSTM (\sname{BiLSTM+GNN}$\rightarrow$ \sname{LSTM}); by extending the Transformer with GNN (\sname{SelfAtt+GNN}$\rightarrow$\sname{SelfAtt}); or by extending their LSTM decoder with a pointer network (\sname{GNN}$\rightarrow$\sname{LSTM+Pointer}). All these extensions can be incorporated into our model as well.


\para{Data Efficiency}
ConvAttention \cite{conv16} performed even better than the Transformer on the Java-small dataset, but could not scale and leverage the larger datasets. Paths+CRFs showed very poor results on the Java-small dataset, which is expected due to the sparse nature of their paths and the CRF model.
When compared to the best among the baselines (BiLSTM with split tokens), our model achieves a relative improvement of $7.3\%$ on Java-large, but as the dataset becomes smaller, the larger the relative difference becomes: $13\%$ on Java-med and $22\%$ on Java-small; when compared to the Transformer, the relative improvement is $23\%$ on Java-large and $37\%$ on Java-small.
These results show the data efficiency of our architecture: while the data-hungry NMT baselines \emph{require} large datasets, our model can leverage both small and large datasets.

\paragraph{Sensitivity to input length} We examined how the performance of each model changes as the size of the test method grows. As shown in \Cref{java-by-length},
our model is superior to all examined baselines across all code lengths.
All models give their best results for short snippets of code, i.e., less than $3$ lines. As the size of the input code increases, all examined models show a natural descent, and show stable results for lengths of $9$ and above.
\definecolor{c2vcolor}{HTML}{F4B400}
\definecolor{treelstmcolor}{HTML}{0F9D58}
\definecolor{bilstmsplitcolor}{HTML}{FF6d00}
\definecolor{transformercolor}{HTML}{46BDC6}
\definecolor{c2scolor}{HTML}{AB30C4}

\begin{figure}[h!]
\hspace{-10mm}
\begin{tikzpicture}[scale=1]
	\begin{axis}[
		xlabel={Code length (lines)},
		ylabel={\footnotesize{F1}},
		ylabel near ticks,
        legend style={at={(1,0.4)},anchor=west,mark size=2pt,font=\scriptsize},
        xmin=0, xmax=31,
        ymin=10, ymax=70,
        xtick={1,2,3,4,5,6,7,8,9,10,11,12,13,14,15,16,17,18,19,20,21,22,23,24,25,26,27,28,29,30},
        xticklabels={1,2,3,4,5,6,7,8,9,10,11,12,13,14,15,16,17,18,19,20,21,22,23,24,25,26,27,28,29,30+},
        ytick={10,15,...,75},
        label style={font=\footnotesize},
        ylabel style={rotate=-90,font=\scriptsize},
        ylabel style={font=\scriptsize},
        xlabel style={font=\footnotesize},
        tick label style={font=\scriptsize} ,
        grid = major,
        major grid style={dotted,black},
        width = \linewidth, height = 5cm
    ]
    \addplot[color=c2scolor,mark options={fill=c2scolor, draw=black, line width=0.5pt}, line width=1pt, mark=*, mark size=2pt] coordinates {
    (1,67.52)
    (2,52.31)
    (3,43.84)
    (4,44.76)
    (5,45.04)
    (6,41.06)
    (7,39.06)
    (8,41.79)
    (9,37.35)
    (10,37.39)
    (11,38.89)
    (12,37.35)
    (13,36.31)
    (14,37.27)
    (15,34.53)
    (16,36.49)
    (17,34.76)
    (18,36.73)
    (19,35.10)
    (20,36.86)
    (21,34.29)
    (22,35.55)
    (23,37.06)
    (24,34.79)
    (25,37.27)
    (26,35.03)
    (27,36.26)
    (28,39.23)
    (29,38.24)
    (30,39.52)
	}; 
    \addlegendentry{\textbf{code2seq} (this work)}

    \addplot[densely dashed,color=bilstmsplitcolor, mark options={solid, fill=bilstmsplitcolor, draw=black}, mark=square*, line width=0.5pt, mark size=2pt] coordinates {
    (1,60.93)
    (2,44.13)
    (3,38.74)
    (4,39.82)
    (5,40.48)
    (6,35.97)
    (7,35.52)
    (8,38.59)
    (9,33.84)
    (10,34.59)
    (11,36.19)
    (12,34.31)
    (13,33.75)
    (14,35.03)
    (15,31.44)
    (16,34.24)
    (17,32.15)
    (18,35.41)
    (19,34.60)
    (20,34.78)
    (21,33.02)
    (22,33.67)
    (23,33.16)
    (24,32.47)
    (25,35.96)
    (26,33.56)
    (27,33.91)
    (28,34.17)
    (29,35.18)
    (30,36.94)
	}; 
    \addlegendentry{2-layer BiLSTMs}

\addplot[densely dotted,color=treelstmcolor, mark options={solid, fill=treelstmcolor, draw=black}, mark=diamond*, line width=0.5pt, mark size=2pt] coordinates {
    (1,60.60)
    (2,44.25)
    (3,37.65)
    (4,38.26)
    (5,38.37)
    (6,34.70)
    (7,31.77)
    (8,34.63)
    (9,31.16)
    (10,31.58)
    (11,33.18)
    (12,31.61)
    (13,30.80)
    (14,31.70)
    (15,29.40)
    (16,30.34)
    (17,28.98)
    (18,31.76)
    (19,29.46)
    (20,31.14)
    (21,29.02)
    (22,30.61)
    (23,30.96)
    (24,30.18)
    (25,30.69)
    (26,30.33)
    (27,27.50)
    (28,29.00)
    (29,32.74)
    (30,38.14 )
	};
    \addlegendentry{TreeLSTM \cite{tai2015improved}}

\addplot[densely dashed,color=transformercolor, mark options={solid, fill=transformercolor, draw=black}, mark=triangle*, line width=0.5pt, mark size=2pt] coordinates {
    (1,54.93)
    (2,36.56)
    (3,32.31)
    (4,32.40)
    (5,33.17)
    (6,29.52)
    (7,27.77)
    (8,31.17)
    (9,27.14)
    (10,28.32)
    (11,29.85)
    (12,28.64)
    (13,27.56)
    (14,28.83)
    (15,26.19)
    (16,28.60)
    (17,26.64)
    (18,28.91)
    (19,27.07)
    (20,29.15)
    (21,26.89)
    (22,28.08)
    (23,28.58)
    (24,27.30)
    (25,29.05)
    (26,27.26)
    (27,26.61)
    (28,27.47)
    (29,29.54)
    (30,31.39)
};
 \addlegendentry{Transformer \cite{vaswani2017attention}}

    \addplot[densely dotted,color=c2vcolor, mark options={solid, fill=c2vcolor, draw=black}, mark=pentagon*, line width=0.5pt, mark size=2pt] coordinates {
    (1,43.49)
    (2,29.99)
    (3,25.46)
    (4,26.40)
    (5,26.89)
    (6,22.63)
    (7,20.84)
    (8,23.47)
    (9,19.75)
    (10,20.67)
    (11,22.31)
    (12,20.65)
    (13,19.49)
    (14,21.25)
    (15,17.70)
    (16,20.52)
    (17,18.53)
    (18,19.78)
    (19,19.27)
    (20,19.84)
    (21,17.84)
    (22,19.30)
    (23,20.45)
    (24,17.07)
    (25,20.37)
    (26,17.83)
    (27,15.71)
    (28,18.14)
    (29,19.85)
    (30,22.46)
	};
    \addlegendentry{code2vec \cite{alon2019code2vec}}

	\end{axis}

\end{tikzpicture}
\caption{F1 score compared to the length of the input code. This experiment was performed for the code summarization task on the Java-med test set. All examples having more than $30$ lines were counted as having $30$ lines.}
\label{java-by-length}
\end{figure}
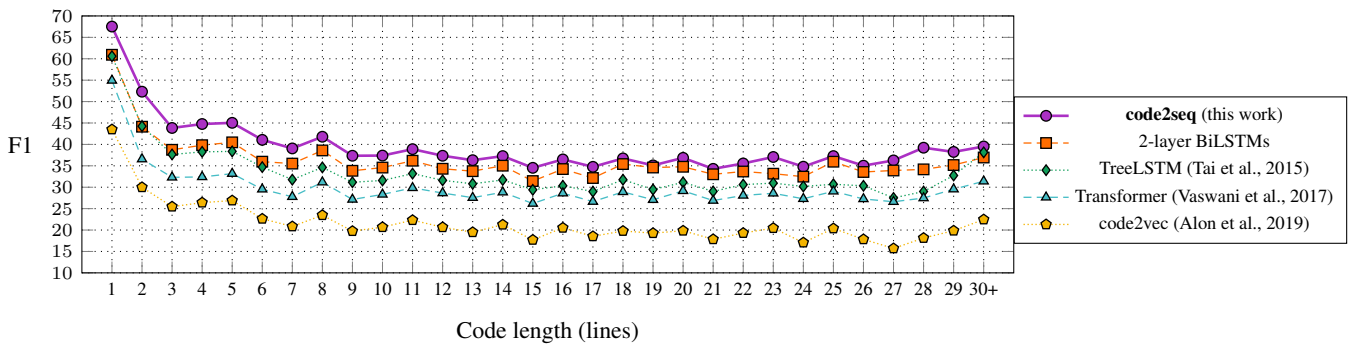 

\subsection{Code Captioning}
\label{subsec:captioning}

For this task we consider predicting a full natural language sentence given a short C\# code snippet. 
We used the dataset of CodeNN \cite{codenn16}, which consists of 66,015 pairs of questions and answers from StackOverflow. They used a semi-supervised classifier to filter irrelevant examples and asked human annotators to provide two additional titles for the examples in the test set, making a total of three reference titles for each code snippet. The target sequence length in this task is about $10$ on average. This dataset is especially challenging as it is orders of magnitude smaller than the code summarization datasets. Additionally, StackOverflow code snippets are typically short, incomplete at times, and aim to provide an answer to a very specific question. We evaluated using BLEU score with smoothing, using the same evaluation scripts as \citet{codenn16}.


\para{Baselines}
We present results compared to CodeNN, TreeLSTMs with attention, 2-layer bidirectional LSTMs with attention, and the Transformer. As before, we provide a fair comparison by splitting tokens to subtokens, and replacing UNK during inference. We also include numbers from baselines used by \citet{codenn16}.

\begin{table}[t]
\caption{Our model outperforms previous work in the code captioning task.  $^{\dag}$Results previously reported by \citet{codenn16}, and verified by us. Another visualization can be found in \Cref{csharp_appendix_section}.} 
\label{csharp-results}
\centering
\footnotesize
\begin{tabular}{ll}
\toprule
Model          & BLEU          \\
\midrule
MOSES$^{\dag}$ \cite{koehn2007moses}          & 11.57          \\
IR$^{\dag}$             & 13.66          \\
SUM-NN$^{\dag}$ \cite{rush2015}         & 19.31          \\
2-layer BiLSTM & 19.78              \\
Transformer \cite{vaswani2017attention}   &  19.68        \\
TreeLSTM \cite{tai2015improved} & 20.11 \\
CodeNN$^{\dag}$ \cite{codenn16}        & 20.53          \\
\midrule 
code2seq & \textbf{23.04} \\
\bottomrule
\end{tabular}
\end{table}

\para{Results}
\Cref{csharp-results} summarizes the results for the code captioning task. Our model achieves a BLEU score of $23.04$, which improves by $2.51$  points (12.2\% relative) over CodeNN, whose authors introduced this dataset, and over all the other baselines, including BiLSTMs, TreeLSTMs and the Transformer, which achieved slightly lower results than CodeNN.
Examples for predictions made by our model and each of the baselines can be found in \Cref{ablation_appendix_section}.
These results show that when the training examples are short and contain incomplete code snippets, our model generalizes better to unseen examples than a shallow textual token-level approach, thanks to its syntactic representation of the data. Although TreeLSTMs also represent the data syntactically, the TreeLSTM baseline achieved lower scores. 

\subsection{Code Documentation}
\label{subsec:documentation}
Although the task of generating code documentation is outside the focus of this work, we performed an additional comparison to \citet{hu2018deep}. They trained a standard seq2seq model by using the linearized AST as the source sequence and a Javadoc natural language sentence as the target sequence. While they originally report a BLEU score of $38.17$, we computed their BLEU score using prediction logs provided us by the authors and obtained a BLEU score of $8.97$, which we find more realistic. Training our model on the same dataset as Hu et al., matching LSTM sizes, and using the same script on our predictions yields a BLEU score of $14.53$, which is a $62\%$ relative gain over the model of \citet{hu2018deep}. This shows that our structural approach represents code better than linearizing the AST and learning it as a sequence.

\section{Ablation Study}
\label{sec:analysis}

\begin{table}[t]
\caption{Variations on the code2seq model, performed on the validation set of Java-med.}
\label{ablation-results}
\centering
\footnotesize
\begin{tabular}{lllll}
\toprule
Model                     & Precision & Recall & F1 & $\Delta$F1\\
\midrule
code2seq (original model) &  \textbf{60.67} & \textbf{47.41}  & \textbf{53.23} & \\
\midrule
No AST nodes (only tokens) & 55.51 & 43.11 & 48.53 & ~~-4.70 \\
No decoder         & 47.99   & 28.96    & 36.12  & -17.11 \\
No token splitting & 48.53 & 34.80 & 40.53 & -12.70 \\
No tokens (only AST nodes) & 33.78 & 21.23 & 26.07 & -27.16 \\
No attention  & 57.00 & 41.89 & 48.29 & ~~-4.94 \\
No random (sample $k$ paths in advance)   & 59.08 & 44.07 & 50.49 & ~~-2.74 \\
\bottomrule
\end{tabular}
\end{table}

To better understand the importance of the different components of our model, we conducted an extensive ablation study. We varied our model in different ways and measured the change in performance. These experiments were performed for the code summarization task, on the validation set of the Java-med dataset. We examined several alternative designs: 


\begin{enumerate}
\item \emph{No AST nodes} -- instead of encoding an AST path using an LSTM, take only the first and last terminal values to construct an input vector 
\item \emph{No decoder} -- no sequential decoding; instead, predict the target sequence as a single symbol using a single softmax layer.
\item \emph{No token splitting} -- no subtoken encoding; instead, embed the full token.
\item \emph{No tokens} -- use only the AST nodes without using the values associated with the terminals.
\item \emph{No attention} -- decode the target sequence given the initial decoder state, without attention.
\item \emph{No random} -- no re-sampling of $k$ paths in each iteration; instead, sample in advance and use the same $k$ paths for each example throughout the training process.
\end{enumerate}


\Cref{ablation-results} shows the results of these alternatives.
As seen, \emph{not encoding AST nodes} resulted in a degradation especially in the precision: a decrease of $5.16$ compared to $4.30$ for the recall. It is quite surprising that this ablation was still better than the baselines (\Cref{methods-results}): for example, the Transformer can implicitly capture pairs of tokens using its self-attention mechanism. However, \emph{not all tokens are AST leaves}. By focusing on AST leaves, we increase the focus on named tokens, and effectively ignore functional tokens like brackets, parentheses, semicolons, etc.
Transformers can (in theory) capture the same signal, but perhaps they require significantly more layers or a different optimization to actually learn to focus on those particular elements. The AST gives us this information for free without having to spend more transformer layers just to learn it.
Additionally, for practical reasons we limited the length of the paths to $9$
. This leads to pairs of leaves that are close in the AST, but not necessarily close in the sequence. In contrast, the Transformer's attention is effectively skewed towards sequential proximity because of the positional embeddings.

Using a single prediction with \emph{no decoder} reduces recall by more than one-third.
This shows that the method name prediction task should be addressed as a sequential prediction, despite the methods' relatively short names.
Using \emph{no token splitting} or \emph{no tokens} at all drastically reduces the score, showing the significance of encoding both subtokens and syntactic paths.
Despite the poor results of \emph{no tokens}, it is still surprising that the model can achieve around half the score of the full model, as using no tokens is equivalent to reasoning about code which has no identifier names, types, APIs, and constant values, which can be very difficult even for a human.
The \emph{no attention} experiment shows the contribution of attention in our model, which is very close in its relative value to the contribution of attention in seq2seq models \cite{luong15, bahdanau14}.
The \emph{no random} experiment shows the positive contribution of sampling $k$ different paths afresh on every training iteration, instead of using the same sample of paths from each example during the entire training. This approach provides data-level regularization that further improves an already powerful model. Another visualization can be found in \Cref{csharp_appendix_section}.

\section{Related Work}

The growing availability of open source repositories creates new opportunities for using machine learning to process source code en masse. Several papers model code as a sequence of tokens \cite{codenn16, conv16, loyola2017}, characters \cite{Bielik2017ProgramSF}, and API calls \cite{raychev14}.
While sometimes obtaining satisfying results, these models treat code as a sequence rather than a tree. This necessitates implicit relearning of the (predefined) syntax of the programming language, wasting resources and reducing accuracy.

Code representation models that use syntactic information have usually been evaluated on relatively easier tasks, which mainly focus on ``filling the blanks'' in a given program \cite{pigeon, phog16, decisionTrees2016, jsnice2015, allamanis2018learning} or semantic classification of code snippets \cite{alon2019code2vec}. Moreover, none of the models that use syntactic relations are compositional, and therefore the number of possible syntactic relations is fixed either before or after training, a process which results in a large RAM and GPU memory consumption.
The syntactic paths of \citet{pigeon, alon2019code2vec} are represented \emph{monolithically}, and are therefore limited to only a subset of the paths that were observed enough times during training. As a result, \emph{they cannot represent unseen relations}. In contrast, by representing AST paths node-by-node using LSTMs, our model can represent and use \emph{any} syntactic path in any unseen example. Further, our model decodes the output sequence step-by-step while attending over the input paths, and can thus generate unseen sequences, compared to code2vec \cite{alon2019code2vec}, which has a closed vocabulary.

\citet{oda2015learning} were the first to generate sequences by leveraging the syntax of code. They performed a line-by-line statistical machine translation (SMT) to translate Python code to pseudo-code. Our tasks are different, and we cannot assume an alignment between elements in the input and the output; our tasks take a whole code snippet as their input, and produce a much shorter sequence as output.
Additionally, a conceptual advantage of our model over line-by-line translation is its ability to capture multiline patterns in the source code. These multiline patterns are often very useful for the model and get the most attention (\Cref{java_example}).
A recent work \cite{hu2018deep} generates comments from code. There is a conceptual difference between our approaches: \citet{hu2018deep} linearize the AST, and then pass it on to a standard seq2seq model. We present a new model, in which the encoder already assumes that the input is tree-structured. When training our model on their dataset, we improve over their BLEU score by $62\%$ (\Cref{subsec:documentation}).

\citet{allamanis2018learning} represent code with Gated Graph Neural Networks. Nodes in the graph represent identifiers, and edges represent syntactic and semantic relations in the code such as ``ComputedFrom'' and ``LastWrite''. The edges are designed for the semantics of a specific programming language, for a specific task, and require an expert to devise and implement. In contrast, our model has minimal assumptions on the input language and is general enough not to require either expert semantic knowledge or the manual design of features. Our model can therefore be easily implemented for various input languages. \citet{bastings2017graph} used graph-convolutional networks for machine translation of natural languages.
\citet{piech2015learning} encoded code using Tree-RNNs to propagate feedback on student code; and \citet{chen2018tree} used Tree-RNNs for a tree-to-tree translation of code into another programming language. 
\section{Conclusion}\label{Conclusion}
We presented a novel code-to-sequence model which considers the unique syntactic structure of source code with a sequential modeling of natural language.
The core idea is to sample paths in the Abstract Syntax Tree of a code snippet, encode these paths with an LSTM, and attend to them while generating the target sequence.

We demonstrate our approach by using it to predict method names across three datasets of varying sizes, predict natural language captions given partial and short code snippets, and to generate method documentation, in two programming languages. Our model performs significantly better than previous programming-language-oriented works and state-of-the-art NMT models applied in our settings.

We believe that the principles presented in this paper can serve as a basis for a wide range of tasks which involve source code and natural language, and can be extended to other kinds of generated outputs. To this end, we make all our code, datasets, and trained models publicly available.

\subsubsection*{Acknowledgments}

We would like to thank Guy Waldman for developing the code2seq website (\url{http://code2seq.org}), Srinivasan Iyer for the guidance in using his C\# dataset, and Miltiadis Allamanis, Yoav Goldberg, Charles Sutton and the anonymous reviewers for their fruitful comments and suggestions.

The research leading to these results has received funding from the European Union's Seventh Framework Programme (FP7) under grant agreement no. 615688-ERC- COG-PRIME. 

\bibliography{bib}

\begin{thebibliography}{41}
\providecommand{\natexlab}[1]{#1}
\providecommand{\url}[1]{\texttt{#1}}
\expandafter\ifx\csname urlstyle\endcsname\relax
  \providecommand{\doi}[1]{doi: #1}\else
  \providecommand{\doi}{doi: \begingroup \urlstyle{rm}\Url}\fi

\bibitem[Allamanis et~al.(2015{\natexlab{a}})Allamanis, Barr, Bird, and
  Sutton]{allamanis2015}
Miltiadis Allamanis, Earl~T. Barr, Christian Bird, and Charles Sutton.
\newblock Suggesting accurate method and class names.
\newblock In \emph{Proceedings of the 2015 10th Joint Meeting on Foundations of
  Software Engineering}, ESEC/FSE 2015, pages 38--49, New York, NY, USA,
  2015{\natexlab{a}}. ACM.
\newblock ISBN 978-1-4503-3675-8.
\newblock \doi{10.1145/2786805.2786849}.
\newblock URL \url{http://doi.acm.org/10.1145/2786805.2786849}.

\bibitem[Allamanis et~al.(2015{\natexlab{b}})Allamanis, Tarlow, Gordon, and
  Wei]{bimodal15}
Miltiadis Allamanis, Daniel Tarlow, Andrew~D. Gordon, and Yi~Wei.
\newblock {Bimodal Modelling of Source Code and Natural Language}.
\newblock In \emph{{Proceedings of the 32nd International Conference on Machine
  Learning}}, volume~37 of \emph{{JMLR Proceedings}}, pages 2123--2132.
  {JMLR.org}, 2015{\natexlab{b}}.

\bibitem[Allamanis et~al.(2016)Allamanis, Peng, and Sutton]{conv16}
Miltiadis Allamanis, Hao Peng, and Charles~A. Sutton.
\newblock A convolutional attention network for extreme summarization of source
  code.
\newblock In \emph{Proceedings of the 33nd International Conference on Machine
  Learning, {ICML} 2016, New York City, NY, USA, June 19-24, 2016}, pages
  2091--2100, 2016.
\newblock URL \url{http://jmlr.org/proceedings/papers/v48/allamanis16.html}.

\bibitem[Allamanis et~al.(2018)Allamanis, Brockschmidt, and
  Khademi]{allamanis2018learning}
Miltiadis Allamanis, Marc Brockschmidt, and Mahmoud Khademi.
\newblock Learning to represent programs with graphs.
\newblock In \emph{International Conference on Learning Representations}, 2018.
\newblock URL \url{https://openreview.net/forum?id=BJOFETxR-}.

\bibitem[Alon et~al.(2018)Alon, Zilberstein, Levy, and Yahav]{pigeon}
Uri Alon, Meital Zilberstein, Omer Levy, and Eran Yahav.
\newblock A general path-based representation for predicting program
  properties.
\newblock In \emph{Proceedings of the 39th ACM SIGPLAN Conference on
  Programming Language Design and Implementation}, PLDI 2018, pages 404--419,
  New York, NY, USA, 2018. ACM.
\newblock ISBN 978-1-4503-5698-5.
\newblock \doi{10.1145/3192366.3192412}.
\newblock URL \url{http://doi.acm.org/10.1145/3192366.3192412}.

\bibitem[Alon et~al.(2019)Alon, Zilberstein, Levy, and Yahav]{alon2019code2vec}
Uri Alon, Meital Zilberstein, Omer Levy, and Eran Yahav.
\newblock Code2vec: Learning distributed representations of code.
\newblock \emph{Proc. ACM Program. Lang.}, 3\penalty0 (POPL):\penalty0
  40:1--40:29, January 2019.
\newblock ISSN 2475-1421.
\newblock \doi{10.1145/3290353}.
\newblock URL \url{http://doi.acm.org/10.1145/3290353}.

\bibitem[Bahdanau et~al.(2014)Bahdanau, Cho, and Bengio]{bahdanau14}
Dzmitry Bahdanau, Kyunghyun Cho, and Yoshua Bengio.
\newblock Neural machine translation by jointly learning to align and
  translate.
\newblock \emph{CoRR}, abs/1409.0473, 2014.
\newblock URL \url{http://arxiv.org/abs/1409.0473}.

\bibitem[Balog et~al.(2017)Balog, Gaunt, Brockschmidt, Nowozin, and
  Tarlow]{balog2016deepcoder}
Matej Balog, Alexander~L Gaunt, Marc Brockschmidt, Sebastian Nowozin, and
  Daniel Tarlow.
\newblock Deepcoder: Learning to write programs.
\newblock In \emph{ICLR}, 2017.

\bibitem[Bastings et~al.(2017)Bastings, Titov, Aziz, Marcheggiani, and
  Simaan]{bastings2017graph}
Joost Bastings, Ivan Titov, Wilker Aziz, Diego Marcheggiani, and Khalil Simaan.
\newblock Graph convolutional encoders for syntax-aware neural machine
  translation.
\newblock In \emph{Proceedings of the 2017 Conference on Empirical Methods in
  Natural Language Processing}, pages 1957--1967, Copenhagen, Denmark,
  September 2017. Association for Computational Linguistics.
\newblock URL \url{https://www.aclweb.org/anthology/D17-1209}.

\bibitem[Bielik et~al.(2016)Bielik, Raychev, and Vechev]{phog16}
Pavol Bielik, Veselin Raychev, and Martin~T. Vechev.
\newblock {PHOG:} probabilistic model for code.
\newblock In \emph{Proceedings of the 33nd International Conference on Machine
  Learning, {ICML} 2016, New York City, NY, USA, June 19-24, 2016}, pages
  2933--2942, 2016.
\newblock URL \url{http://jmlr.org/proceedings/papers/v48/bielik16.html}.

\bibitem[Bielik et~al.(2017)Bielik, Raychev, and Vechev]{Bielik2017ProgramSF}
Pavol Bielik, Veselin Raychev, and Martin Vechev.
\newblock Program synthesis for character level language modeling.
\newblock In \emph{ICLR}, 2017.

\bibitem[Brockschmidt et~al.(2019)Brockschmidt, Allamanis, Gaunt, and
  Polozov]{brockschmidt2018generative}
Marc Brockschmidt, Miltiadis Allamanis, Alexander~L. Gaunt, and Oleksandr
  Polozov.
\newblock Generative code modeling with graphs.
\newblock In \emph{International Conference on Learning Representations}, 2019.
\newblock URL \url{https://openreview.net/forum?id=Bke4KsA5FX}.

\bibitem[Chen et~al.(2018)Chen, Liu, and Song]{chen2018tree}
Xinyun Chen, Chang Liu, and Dawn Song.
\newblock Tree-to-tree neural networks for program translation.
\newblock \emph{CoRR}, abs/1802.03691, 2018.
\newblock URL \url{http://arxiv.org/abs/1802.03691}.

\bibitem[Cho et~al.(2014)Cho, Van~Merri{\"e}nboer, Gulcehre, Bahdanau,
  Bougares, Schwenk, and Bengio]{cho2014learning}
Kyunghyun Cho, Bart Van~Merri{\"e}nboer, Caglar Gulcehre, Dzmitry Bahdanau,
  Fethi Bougares, Holger Schwenk, and Yoshua Bengio.
\newblock Learning phrase representations using rnn encoder-decoder for
  statistical machine translation.
\newblock \emph{arXiv preprint arXiv:1406.1078}, 2014.

\bibitem[Devlin et~al.(2017)Devlin, Uesato, Bhupatiraju, Singh, Mohamed, and
  Kohli]{devlin2017robustfill}
Jacob Devlin, Jonathan Uesato, Surya Bhupatiraju, Rishabh Singh, Abdel-rahman
  Mohamed, and Pushmeet Kohli.
\newblock Robustfill: Neural program learning under noisy i/o.
\newblock In \emph{International Conference on Machine Learning}, pages
  990--998, 2017.

\bibitem[Fernandes et~al.(2019)Fernandes, Allamanis, and
  Brockschmidt]{fernandes2018structured}
Patrick Fernandes, Miltiadis Allamanis, and Marc Brockschmidt.
\newblock Structured neural summarization.
\newblock In \emph{International Conference on Learning Representations}, 2019.
\newblock URL \url{https://openreview.net/forum?id=H1ersoRqtm}.

\bibitem[Glorot and Bengio(2010)]{glorot2010understanding}
Xavier Glorot and Yoshua Bengio.
\newblock Understanding the difficulty of training deep feedforward neural
  networks.
\newblock In \emph{Proceedings of the Thirteenth International Conference on
  Artificial Intelligence and Statistics}, pages 249--256, 2010.

\bibitem[Hochreiter and Schmidhuber(1997)]{hochreiter1997lstm}
Sepp Hochreiter and J\"{u}rgen Schmidhuber.
\newblock Long short-term memory.
\newblock \emph{Neural Comput.}, 9\penalty0 (8):\penalty0 1735--1780, November
  1997.
\newblock ISSN 0899-7667.
\newblock \doi{10.1162/neco.1997.9.8.1735}.
\newblock URL \url{http://dx.doi.org/10.1162/neco.1997.9.8.1735}.

\bibitem[Hu et~al.(2018)Hu, Li, Xia, Lo, and Jin]{hu2018deep}
Xing Hu, Ge~Li, Xin Xia, David Lo, and Zhi Jin.
\newblock Deep code comment generation.
\newblock In \emph{Proceedings of the 26th Conference on Program
  Comprehension}, pages 200--210. ACM, 2018.

\bibitem[Iyer et~al.(2016)Iyer, Konstas, Cheung, and Zettlemoyer]{codenn16}
Srinivasan Iyer, Ioannis Konstas, Alvin Cheung, and Luke Zettlemoyer.
\newblock Summarizing source code using a neural attention model.
\newblock In \emph{Proceedings of the 54th Annual Meeting of the Association
  for Computational Linguistics, {ACL} 2016, August 7-12, 2016, Berlin,
  Germany, Volume 1: Long Papers}, 2016.
\newblock URL \url{http://aclweb.org/anthology/P/P16/P16-1195.pdf}.

\bibitem[{Klein} et~al.(2017){Klein}, {Kim}, {Deng}, {Senellart}, and
  {Rush}]{2017opennmt}
G.~{Klein}, Y.~{Kim}, Y.~{Deng}, J.~{Senellart}, and A.~M. {Rush}.
\newblock {OpenNMT: Open-Source Toolkit for Neural Machine Translation}.
\newblock \emph{ArXiv e-prints}, 2017.

\bibitem[Koehn et~al.(2007)Koehn, Hoang, Birch, Callison-Burch, Federico,
  Bertoldi, Cowan, Shen, Moran, Zens, Dyer, Bojar, Constantin, and
  Herbst]{koehn2007moses}
Philipp Koehn, Hieu Hoang, Alexandra Birch, Chris Callison-Burch, Marcello
  Federico, Nicola Bertoldi, Brooke Cowan, Wade Shen, Christine Moran, Richard
  Zens, Chris Dyer, Ond\v{r}ej Bojar, Alexandra Constantin, and Evan Herbst.
\newblock Moses: Open source toolkit for statistical machine translation.
\newblock In \emph{Proceedings of the 45th Annual Meeting of the ACL on
  Interactive Poster and Demonstration Sessions}, ACL '07, pages 177--180,
  Stroudsburg, PA, USA, 2007. Association for Computational Linguistics.
\newblock URL \url{http://dl.acm.org/citation.cfm?id=1557769.1557821}.

\bibitem[Loyola et~al.(2017)Loyola, Marrese-Taylor, and Matsuo]{loyola2017}
Pablo Loyola, Edison Marrese-Taylor, and Yutaka Matsuo.
\newblock A neural architecture for generating natural language descriptions
  from source code changes.
\newblock In \emph{Proceedings of the 55th Annual Meeting of the Association
  for Computational Linguistics (Volume 2: Short Papers)}, pages 287--292.
  Association for Computational Linguistics, 2017.
\newblock \doi{10.18653/v1/P17-2045}.
\newblock URL \url{http://www.aclweb.org/anthology/P17-2045}.

\bibitem[Luong et~al.(2015)Luong, Pham, and Manning]{luong15}
Thang Luong, Hieu Pham, and Christopher~D. Manning.
\newblock Effective approaches to attention-based neural machine translation.
\newblock In \emph{Proceedings of the 2015 Conference on Empirical Methods in
  Natural Language Processing, {EMNLP} 2015, Lisbon, Portugal, September 17-21,
  2015}, pages 1412--1421, 2015.
\newblock URL \url{http://aclweb.org/anthology/D/D15/D15-1166.pdf}.

\bibitem[Murali et~al.(2017)Murali, Chaudhuri, and Jermaine]{murali2018bayou}
Vijayaraghavan Murali, Swarat Chaudhuri, and Chris Jermaine.
\newblock Bayesian sketch learning for program synthesis.
\newblock \emph{CoRR}, abs/1703.05698, 2017.
\newblock URL \url{http://arxiv.org/abs/1703.05698}.

\bibitem[Nesterov(1983)]{nesterov1983method}
Yurii~E Nesterov.
\newblock A method for solving the convex programming problem with convergence
  rate o (1/k\^{} 2).
\newblock In \emph{Dokl. Akad. Nauk SSSR}, volume 269, pages 543--547, 1983.

\bibitem[Oda et~al.(2015)Oda, Fudaba, Neubig, Hata, Sakti, Toda, and
  Nakamura]{oda2015learning}
Yusuke Oda, Hiroyuki Fudaba, Graham Neubig, Hideaki Hata, Sakriani Sakti,
  Tomoki Toda, and Satoshi Nakamura.
\newblock Learning to generate pseudo-code from source code using statistical
  machine translation (t).
\newblock In \emph{Automated Software Engineering (ASE), 2015 30th IEEE/ACM
  International Conference on}, pages 574--584. IEEE, 2015.

\bibitem[Piech et~al.(2015)Piech, Huang, Nguyen, Phulsuksombati, Sahami, and
  Guibas]{piech2015learning}
Chris Piech, Jonathan Huang, Andy Nguyen, Mike Phulsuksombati, Mehran Sahami,
  and Leonidas Guibas.
\newblock Learning program embeddings to propagate feedback on student code.
\newblock In \emph{Proceedings of the 32Nd International Conference on
  International Conference on Machine Learning - Volume 37}, ICML'15, pages
  1093--1102. JMLR.org, 2015.
\newblock URL \url{http://dl.acm.org/citation.cfm?id=3045118.3045235}.

\bibitem[Rabinovich et~al.(2017)Rabinovich, Stern, and Klein]{rabinovich2017}
Maxim Rabinovich, Mitchell Stern, and Dan Klein.
\newblock Abstract syntax networks for code generation and semantic parsing.
\newblock In \emph{Proceedings of the 55th Annual Meeting of the Association
  for Computational Linguistics (Volume 1: Long Papers)}, pages 1139--1149.
  Association for Computational Linguistics, 2017.
\newblock \doi{10.18653/v1/P17-1105}.
\newblock URL \url{http://www.aclweb.org/anthology/P17-1105}.

\bibitem[Raychev et~al.(2014)Raychev, Vechev, and Yahav]{raychev14}
Veselin Raychev, Martin Vechev, and Eran Yahav.
\newblock Code completion with statistical language models.
\newblock \emph{SIGPLAN Not.}, 49\penalty0 (6):\penalty0 419--428, June 2014.
\newblock ISSN 0362-1340.
\newblock \doi{10.1145/2666356.2594321}.
\newblock URL \url{http://doi.acm.org/10.1145/2666356.2594321}.

\bibitem[Raychev et~al.(2015)Raychev, Vechev, and Krause]{jsnice2015}
Veselin Raychev, Martin Vechev, and Andreas Krause.
\newblock Predicting program properties from "big code".
\newblock In \emph{Proceedings of the 42Nd Annual ACM SIGPLAN-SIGACT Symposium
  on Principles of Programming Languages}, POPL '15, pages 111--124, New York,
  NY, USA, 2015. ACM.
\newblock ISBN 978-1-4503-3300-9.
\newblock \doi{10.1145/2676726.2677009}.
\newblock URL \url{http://doi.acm.org/10.1145/2676726.2677009}.

\bibitem[Raychev et~al.(2016)Raychev, Bielik, and Vechev]{decisionTrees2016}
Veselin Raychev, Pavol Bielik, and Martin Vechev.
\newblock Probabilistic model for code with decision trees.
\newblock In \emph{Proceedings of the 2016 ACM SIGPLAN International Conference
  on Object-Oriented Programming, Systems, Languages, and Applications}, OOPSLA
  2016, pages 731--747, New York, NY, USA, 2016. ACM.
\newblock ISBN 978-1-4503-4444-9.
\newblock \doi{10.1145/2983990.2984041}.
\newblock URL \url{http://doi.acm.org/10.1145/2983990.2984041}.

\bibitem[Rubinstein(1999)]{rubinstein1999cross}
Reuven Rubinstein.
\newblock The cross-entropy method for combinatorial and continuous
  optimization.
\newblock \emph{Methodology and Computing in Applied Probability}, 1\penalty0
  (2):\penalty0 127--190, 1999.

\bibitem[Rubinstein(2001)]{rubinstein2001combinatorial}
Reuven~Y Rubinstein.
\newblock Combinatorial optimization, cross-entropy, ants and rare events.
\newblock \emph{Stochastic Optimization: Algorithms and Applications},
  54:\penalty0 303--363, 2001.

\bibitem[Rush et~al.(2015)Rush, Chopra, and Weston]{rush2015}
Alexander~M. Rush, Sumit Chopra, and Jason Weston.
\newblock A neural attention model for abstractive sentence summarization.
\newblock In \emph{Proceedings of the 2015 Conference on Empirical Methods in
  Natural Language Processing, {EMNLP} 2015, Lisbon, Portugal, September 17-21,
  2015}, pages 379--389, 2015.
\newblock URL \url{http://aclweb.org/anthology/D/D15/D15-1044.pdf}.

\bibitem[Sennrich et~al.(2016)Sennrich, Haddow, and Birch]{sennrich2016}
Rico Sennrich, Barry Haddow, and Alexandra Birch.
\newblock Neural machine translation of rare words with subword units.
\newblock In \emph{Proceedings of the 54th Annual Meeting of the Association
  for Computational Linguistics (Volume 1: Long Papers)}, pages 1715--1725,
  Berlin, Germany, August 2016. Association for Computational Linguistics.
\newblock URL \url{http://www.aclweb.org/anthology/P16-1162}.

\bibitem[Srivastava et~al.(2014)Srivastava, Hinton, Krizhevsky, Sutskever, and
  Salakhutdinov]{srivastava2014dropout}
Nitish Srivastava, Geoffrey~E Hinton, Alex Krizhevsky, Ilya Sutskever, and
  Ruslan Salakhutdinov.
\newblock Dropout: a simple way to prevent neural networks from overfitting.
\newblock \emph{Journal of Machine Learning Research}, 15\penalty0
  (1):\penalty0 1929--1958, 2014.

\bibitem[Sutskever et~al.(2014)Sutskever, Vinyals, and
  Le]{sutskever2014sequence}
Ilya Sutskever, Oriol Vinyals, and Quoc~V Le.
\newblock Sequence to sequence learning with neural networks.
\newblock In \emph{Advances in Neural Information Processing Systems}, pages
  3104--3112, 2014.

\bibitem[Tai et~al.(2015)Tai, Socher, and Manning]{tai2015improved}
Kai~Sheng Tai, Richard Socher, and Christopher~D. Manning.
\newblock Improved semantic representations from tree-structured long
  short-term memory networks.
\newblock In \emph{Proceedings of the 53rd Annual Meeting of the Association
  for Computational Linguistics and the 7th International Joint Conference on
  Natural Language Processing (Volume 1: Long Papers)}, pages 1556--1566.
  Association for Computational Linguistics, 2015.
\newblock \doi{10.3115/v1/P15-1150}.
\newblock URL \url{http://aclweb.org/anthology/P15-1150}.

\bibitem[Vaswani et~al.(2017)Vaswani, Shazeer, Parmar, Uszkoreit, Jones, Gomez,
  Kaiser, and Polosukhin]{vaswani2017attention}
Ashish Vaswani, Noam Shazeer, Niki Parmar, Jakob Uszkoreit, Llion Jones,
  Aidan~N Gomez, {\L}ukasz Kaiser, and Illia Polosukhin.
\newblock Attention is all you need.
\newblock In \emph{Advances in Neural Information Processing Systems}, pages
  6000--6010, 2017.

\bibitem[Yin and Neubig(2017)]{yin2017}
Pengcheng Yin and Graham Neubig.
\newblock A syntactic neural model for general-purpose code generation.
\newblock In \emph{Proceedings of the 55th Annual Meeting of the Association
  for Computational Linguistics (Volume 1: Long Papers)}, pages 440--450.
  Association for Computational Linguistics, 2017.
\newblock \doi{10.18653/v1/P17-1041}.
\newblock URL \url{http://www.aclweb.org/anthology/P17-1041}.

\end{thebibliography}
\bibliographystyle{plainnat}

\clearpage
\appendix

\section{Additional evaluation}\label{appendix_details}
\paragraph{Comparison to code2vec on their dataset}
We perform an additional comparison to code2vec \cite{alon2019code2vec} on their proposed dataset.
As shown in \Cref{results_code2vec_dataset}, code2vec achieves a high F1 score on that dataset. However, our model achieves an even higher F1 score.
The poorer performance of code2vec on our dataset is probably due to its always being split to train/validation/test \emph {by project}, whereas the dataset of code2vec is split \emph {by file}.
In the code2vec dataset, a file can be in the training set, while another file from the same project can be in the test set. This makes their dataset significantly easier, because method names ``leak'' to other files in the same project, and there are often duplicates in different files of the same project. This is consistent with \citet{allamanis2018learning}, who found that splitting by file makes the dataset easier than by project.
We decided to take the stricter approach, and not to use their dataset (even though our model achieves better results on it), in order to make all of our comparisons on split-by-project datasets.

\begin{table}[t]
\caption{Our model significantly outperforms code2vec on the code2vec dataset.}
\label{results_code2vec_dataset}
\scriptsize
\centering
\makebox[\linewidth]{
\begin{tabular}{lllllll}
\toprule
\multirow{2}{*}{Model}           & \multicolumn{3}{c}{Our Java-large dataset (same as in \Cref{methods-results})}                   & \multicolumn{3}{c}{code2vec dataset \cite{alon2019code2vec}}  \\
\cmidrule{2-4} \cmidrule{5-7}
                                 &  Prec      & Rec         & F1   & Prec      & Rec         & F1      \\
\midrule
code2vec \cite{alon2019code2vec} & 48.2 & 38.4 & 42.7 & 63.1 & 54.4 & 58.4 \\
\midrule 
code2seq (this work)                        & \textbf{64.0}            & \textbf{55.02}      & \textbf{59.2} & \textbf{70.2} & \textbf{63.3} & \textbf{66.6}\\
\bottomrule
\end{tabular}
}
\end{table}

\begin{table}[t]
\caption{Statistics of our datasets.}
\label{data_stats_table}
\scriptsize
\centering
\makebox[\linewidth]{
\begin{tabular}{lllll}
\toprule
                                & Java-small    & Java-med    & Java-large & C\# \cite{codenn16}    \\
\midrule
\#projects - training           & 10            & 800         & 8999     & - \\
\#projects - validation         & 1             & 100         & 250      & - \\
\#projects - test               & 1             & 96          & 307      & - \\
\#examples - training           & 665,115       & 3,004,536   & 15,344,512 & 52,812 \\
\#examples - validation         & 23,505        & 410,699     & 320,866    & 6,601 \\
\#examples - test               & 56,165        & 411,751     & 417,003    & 6,602 \\
Avg. number of paths (training) & 171           & 187         & 220      & 207 \\
Avg. code length - lines (training) & 6.0 & 6.3 & 6.6 & 8.3 \\
Avg. code length - tokens (training)      & 60            & 63          & 65       & 38\\
Avg. code length - subtokenized (training) & 75            & 78          & 80       & 67\\
Avg. target length (training)             & 3             & 3           & 3        & 10 \\
\bottomrule
\end{tabular}
}
\end{table} 

\paragraph{Data statistics}
\Cref{data_stats_table} shows some statistics of our used datasets.

\section{Code captioning examples}\label{appendix_csharp}
\Cref{appendix_csharp_figure} contains examples from our test set for the code captioning task in C\#, along with the prediction of our model and each of the baselines.

\Cref{fig:treenode} shows a timestep-by-timestep example, with the symbol decoded at each timestep and the top-attended path at that step. The width of the path is proportional to the attention it was given by the model (because of space limitations we included only the top-attended path for each decoding step, but hundreds of paths are attended at each step).

\section{Code summarization examples}\label{appendix_java}
\Cref{appendix_java_figure} contains examples from our test set for the code summarization task in C\#, along with the prediction of our model and each of the baselines. The presented predictions are made by models that were trained on the same Java-large dataset.

\newcommand{\treenodeheight}{4.2cm}
\newcommand{\treenodefigurewidth}{0.27\textwidth}
\newcommand{\treenodecaptionwidth}{0.23\textwidth}
\newcommand{\raisecaption}{2.5mm}
\begin{figure*}[t]

\centering
\begin{minipage}{\textwidth}
\centering
\begin{tabular}{clcl}
\begin{subfigure}[t]{\treenodefigurewidth}
\includegraphics[height=\treenodeheight,keepaspectratio]{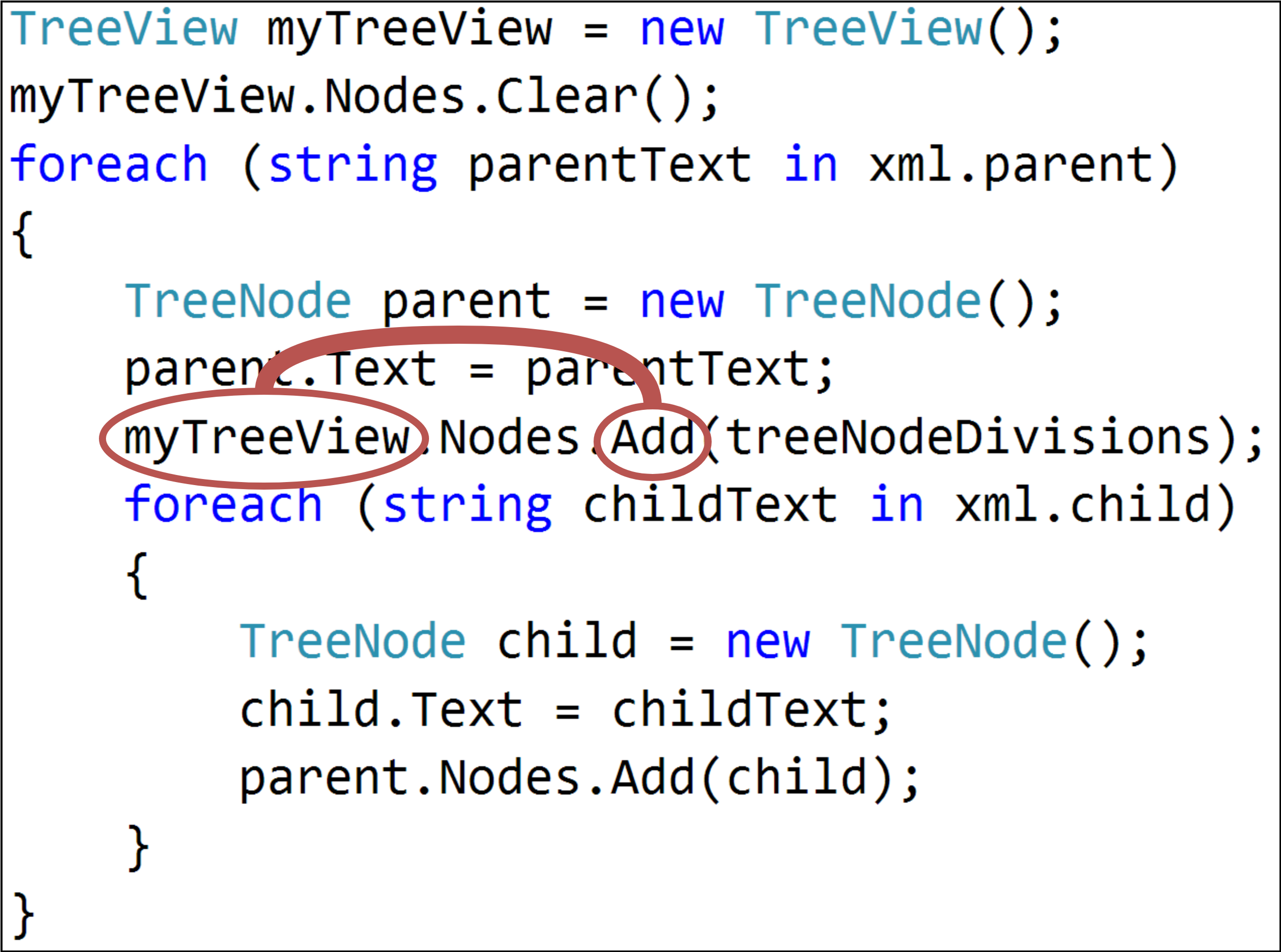}
\end{subfigure}


\begin{subfigure}[t]{\treenodecaptionwidth}
{\raisebox{\raisecaption}{
{\colorbox{white}{\framebox{\large \textbf{add}}}}
}}
\end{subfigure}

\begin{subfigure}[t]{\treenodefigurewidth}

\includegraphics[height=\treenodeheight,keepaspectratio]{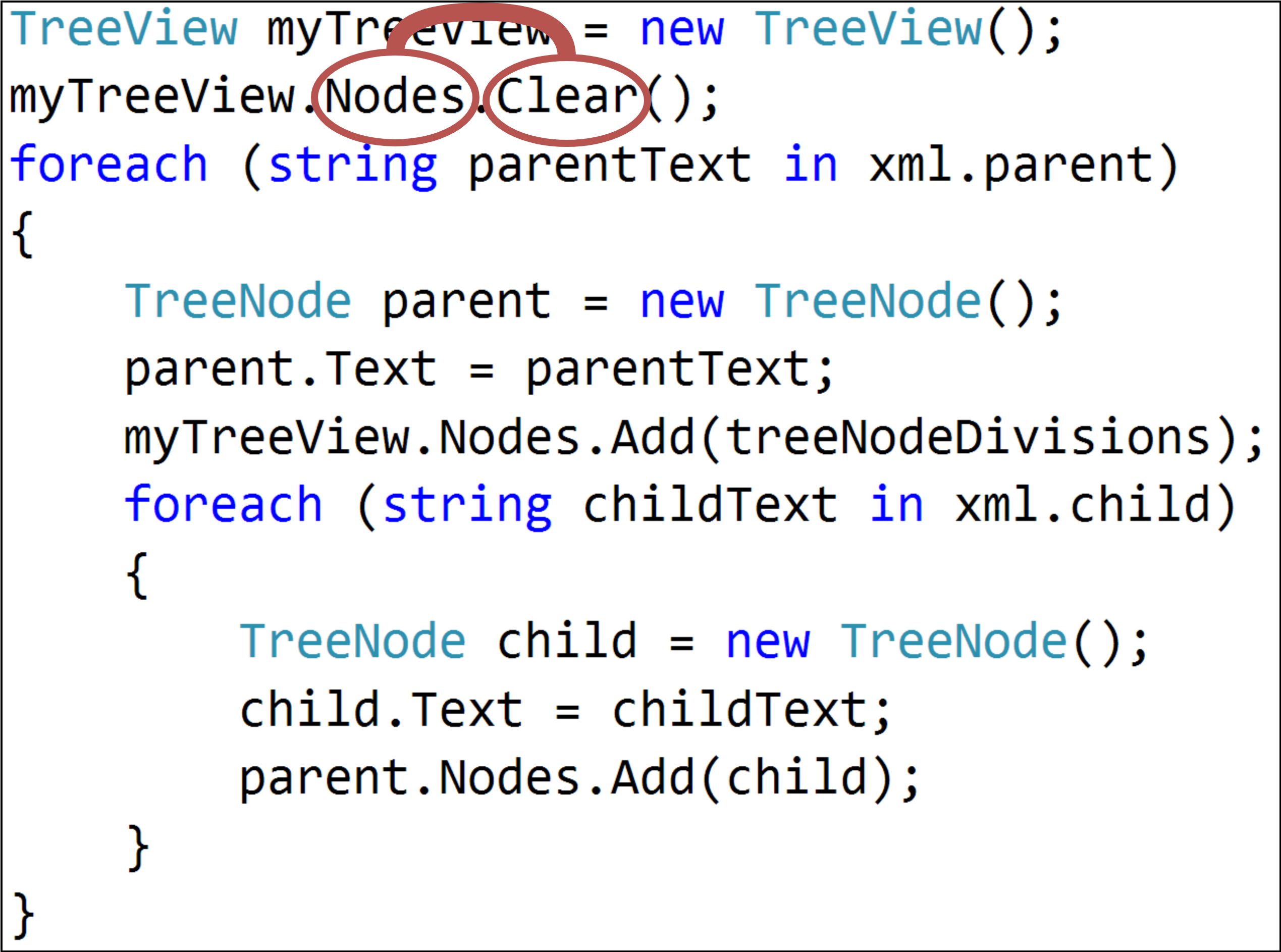}
\end{subfigure}

\begin{subfigure}[t]{\treenodecaptionwidth}
{\raisebox{\raisecaption}{
{\colorbox{white}{\framebox{\large \textbf{a}}}}
}}
\end{subfigure}
\\
\begin{subfigure}[t]{\treenodefigurewidth}

\includegraphics[height=\treenodeheight,keepaspectratio]{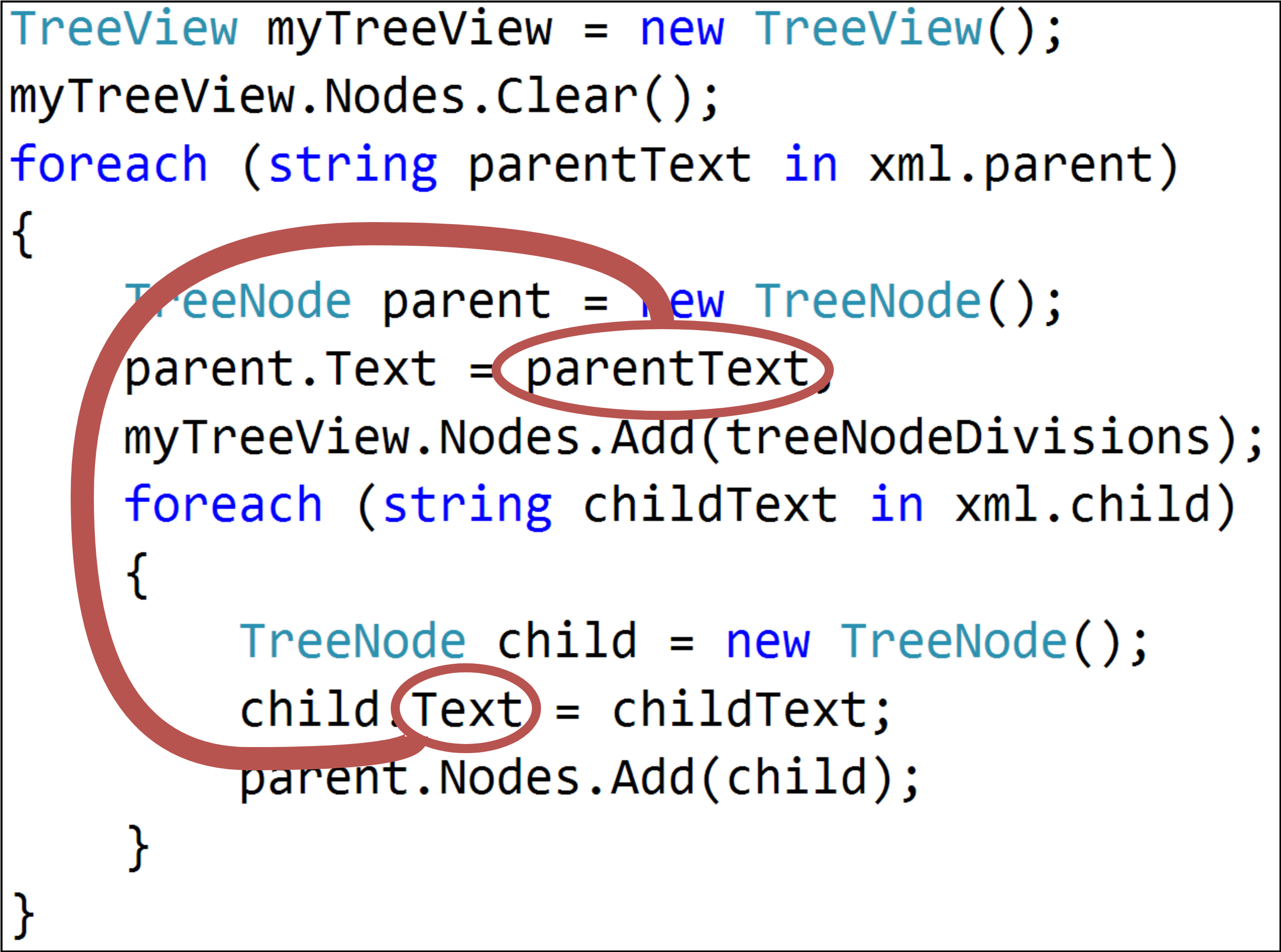}
\end{subfigure}

\begin{subfigure}[t]{\treenodecaptionwidth}
{\raisebox{\raisecaption}{
{\colorbox{white}{\framebox{\large \textbf{child}}}}
}}
\end{subfigure}

\begin{subfigure}[t]{\treenodefigurewidth}

\includegraphics[height=\treenodeheight,keepaspectratio]{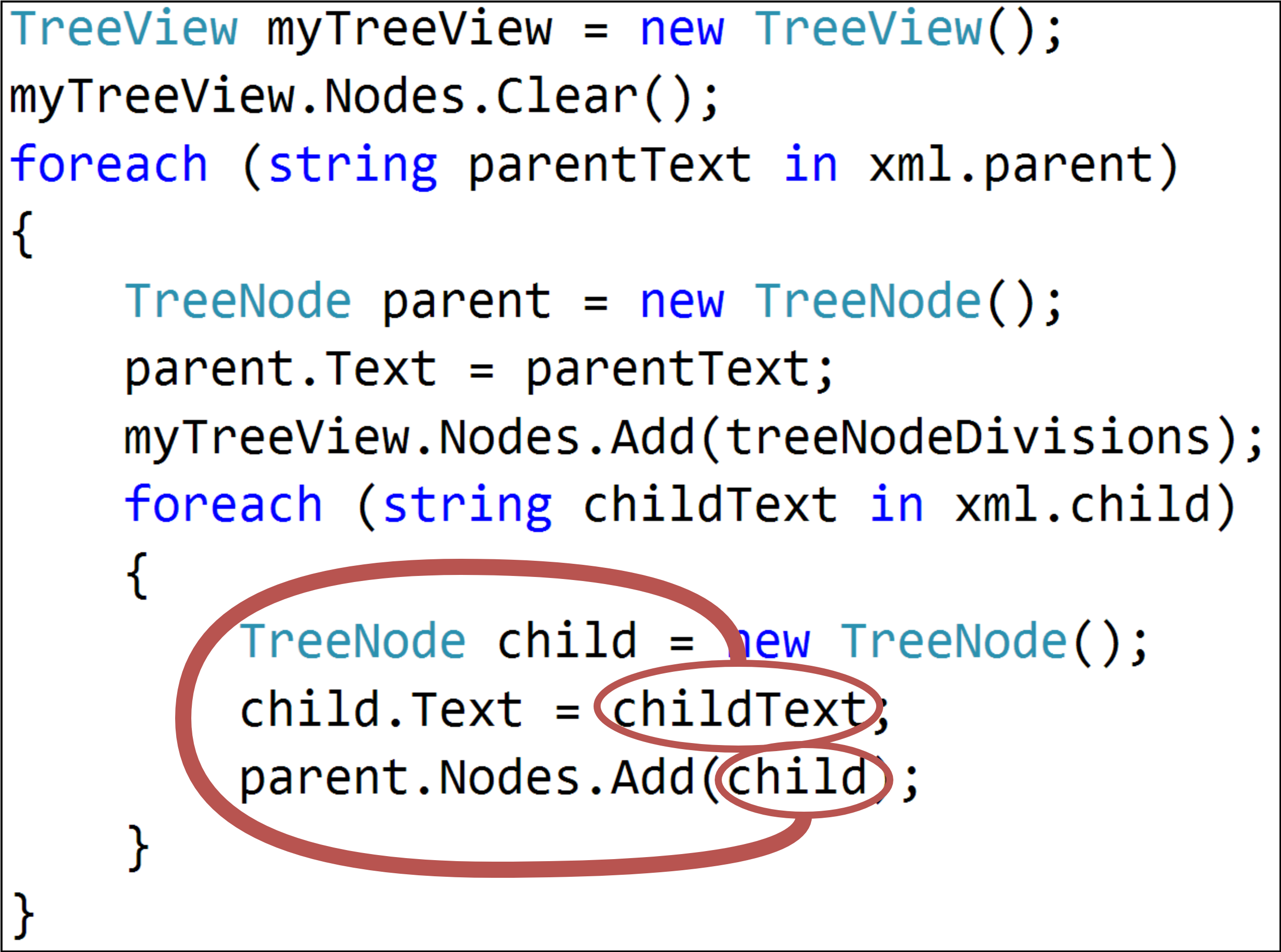}
\end{subfigure}

\begin{subfigure}[t]{\treenodecaptionwidth}
{\raisebox{\raisecaption}{
{\colorbox{white}{\framebox{\large \textbf{node}}}}
}}
\end{subfigure}
\\

\begin{subfigure}[t]{\treenodefigurewidth}

\includegraphics[height=\treenodeheight,keepaspectratio]{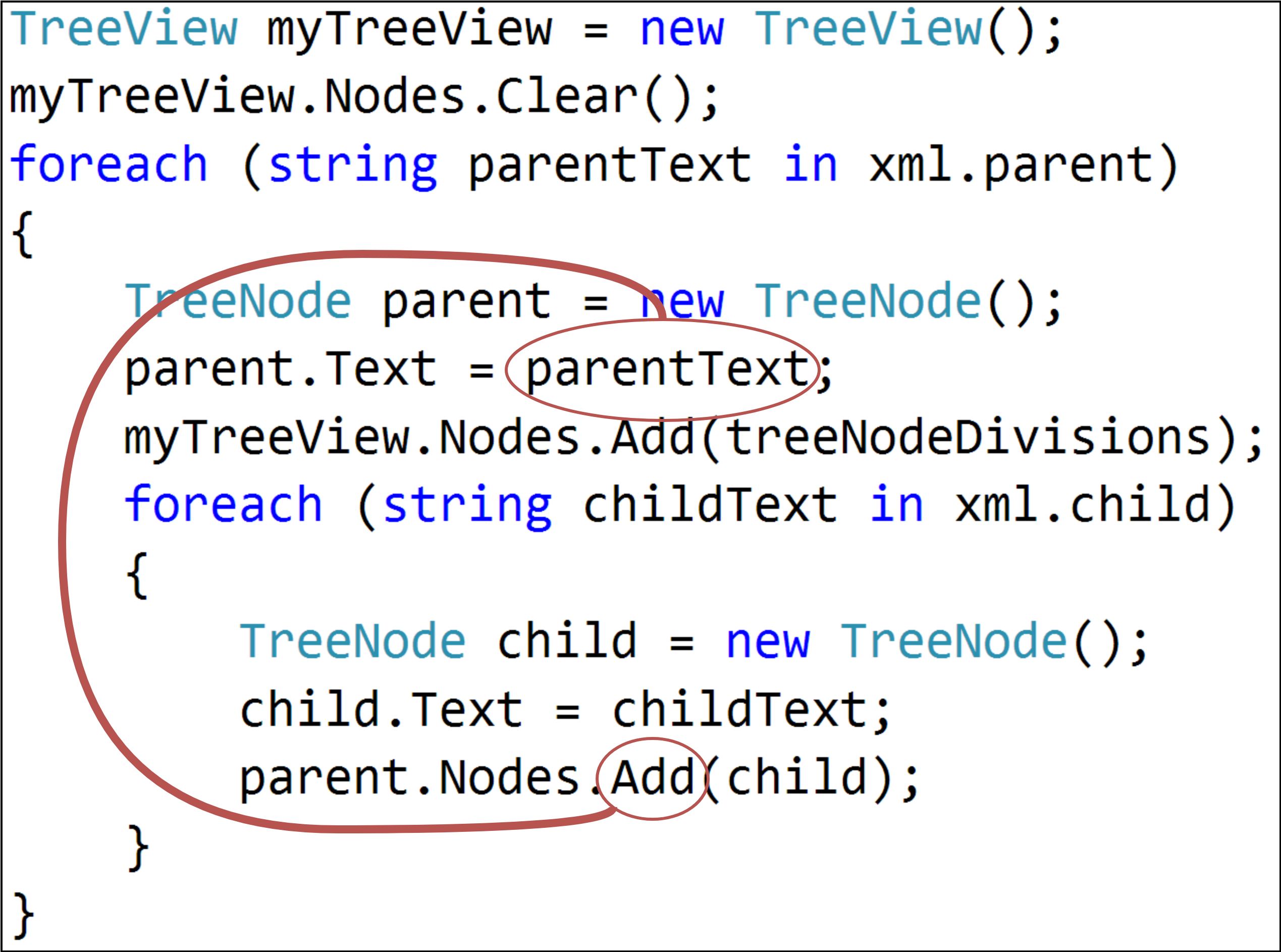}
\end{subfigure}


\begin{subfigure}[t]{\treenodecaptionwidth}
{\raisebox{\raisecaption}{
{\colorbox{white}{\framebox{\large \textbf{to}}}}
}}
\end{subfigure}

\begin{subfigure}[t]{\treenodefigurewidth}

\includegraphics[height=\treenodeheight,keepaspectratio]{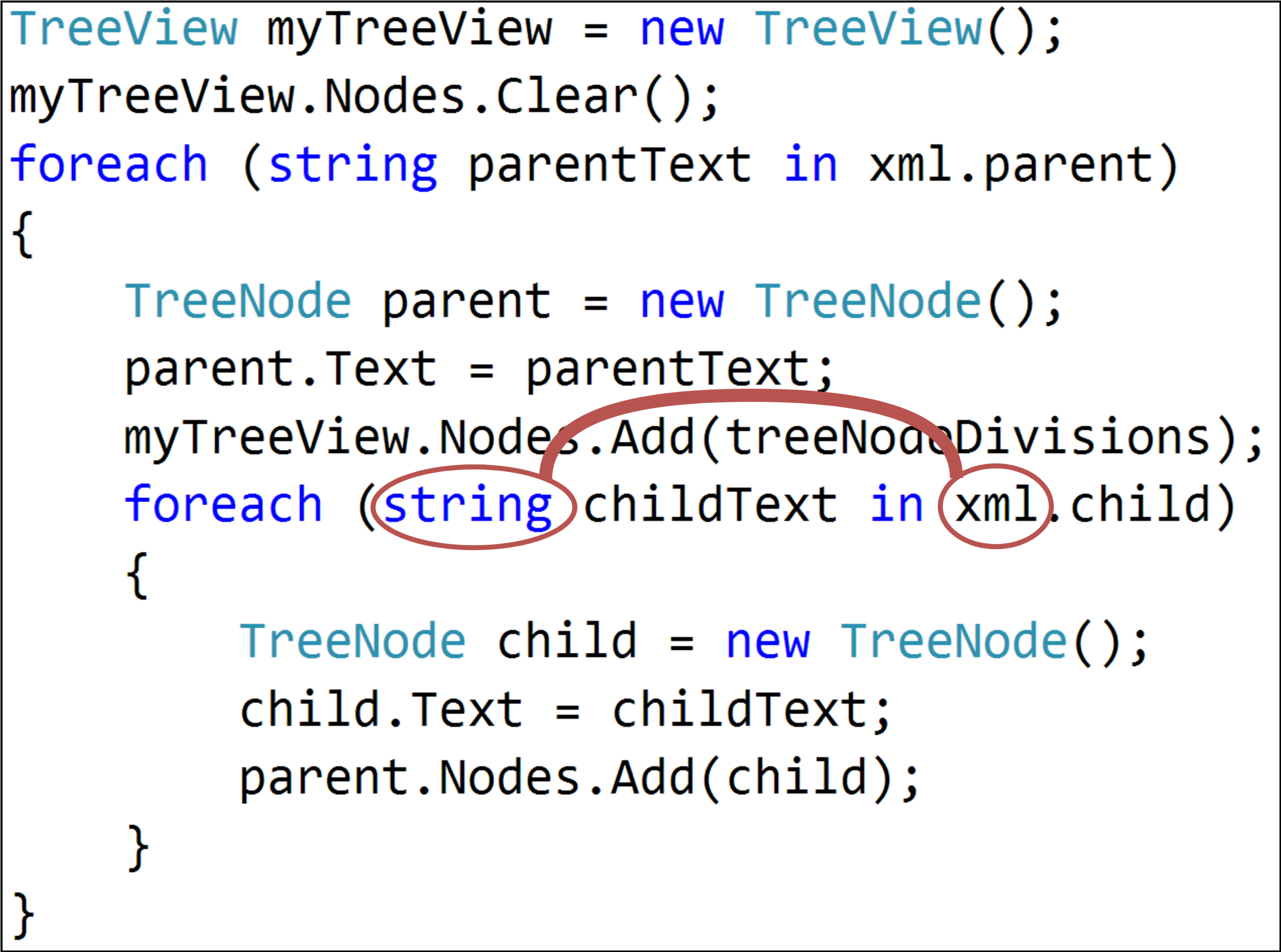}
\end{subfigure}

\begin{subfigure}[t]{\treenodecaptionwidth}
{\raisebox{\raisecaption}{
{\colorbox{white}{\framebox{\large \textbf{a}}}}
}}
\end{subfigure}
\\

\begin{subfigure}[t]{\treenodefigurewidth}

\includegraphics[height=\treenodeheight,keepaspectratio]{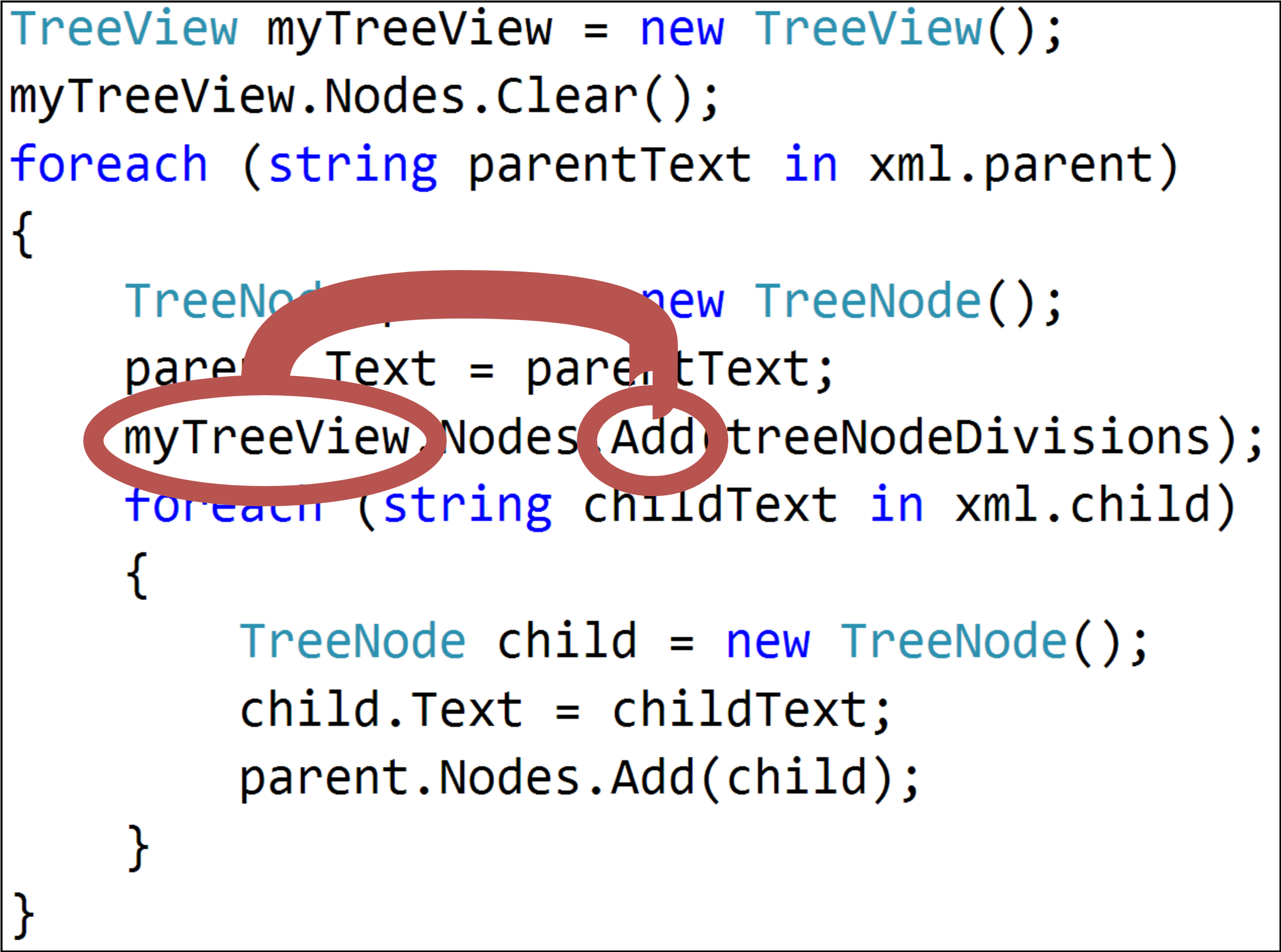}
\end{subfigure}


\begin{subfigure}[t]{\treenodecaptionwidth}
{\raisebox{\raisecaption}{
{\colorbox{white}{\framebox{\large \textbf{treeview}}}}
}}
\end{subfigure}

\begin{subfigure}[t]{\treenodefigurewidth}

\includegraphics[height=\treenodeheight,keepaspectratio]{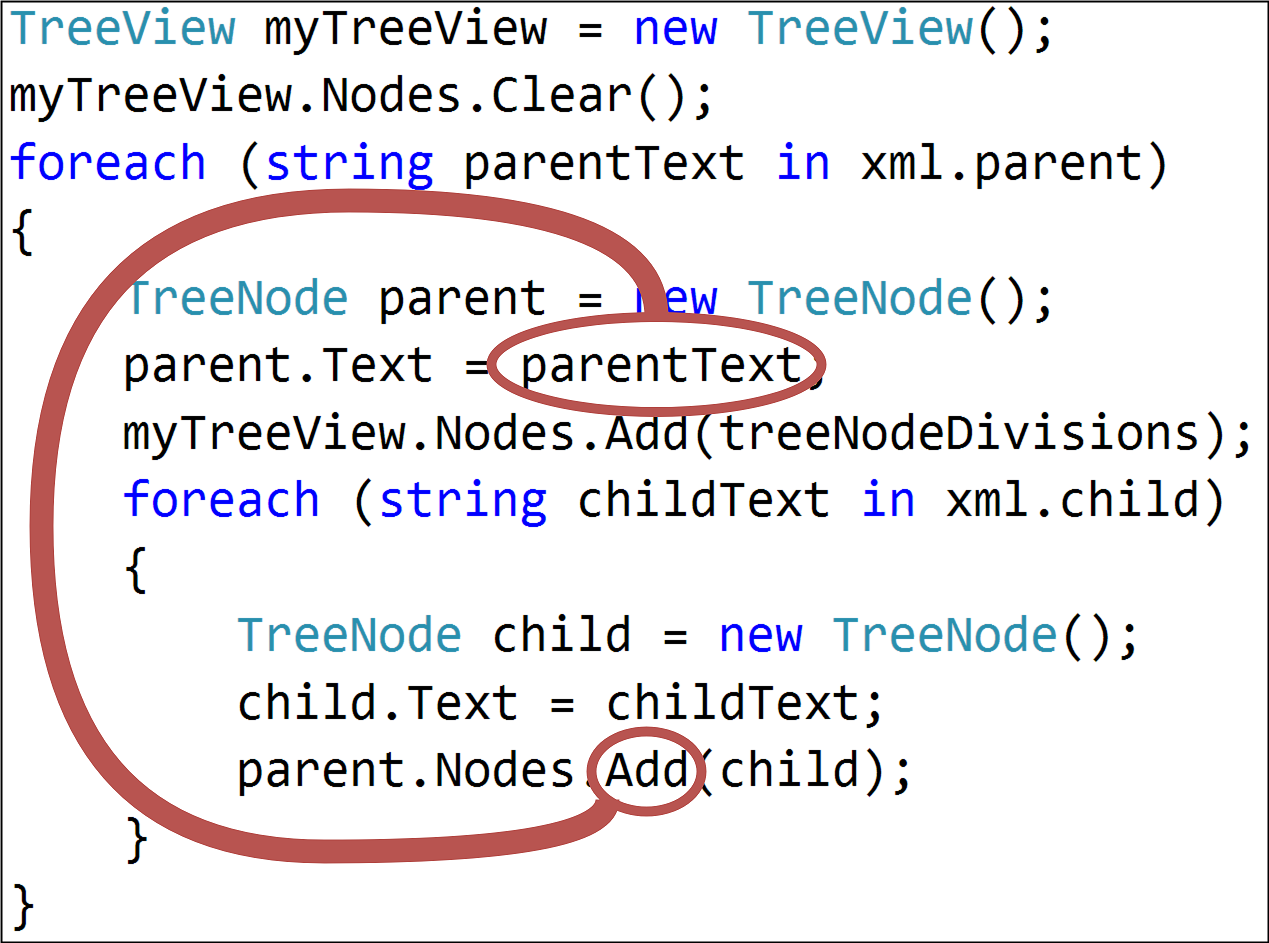}
\end{subfigure}

\begin{subfigure}[t]{\treenodecaptionwidth}
{\raisebox{\raisecaption}{
{\colorbox{white}{\framebox{\large \textbf{in}}}}
}}
\end{subfigure}
\\

\begin{subfigure}[t]{\treenodefigurewidth}

\includegraphics[height=\treenodeheight,keepaspectratio]{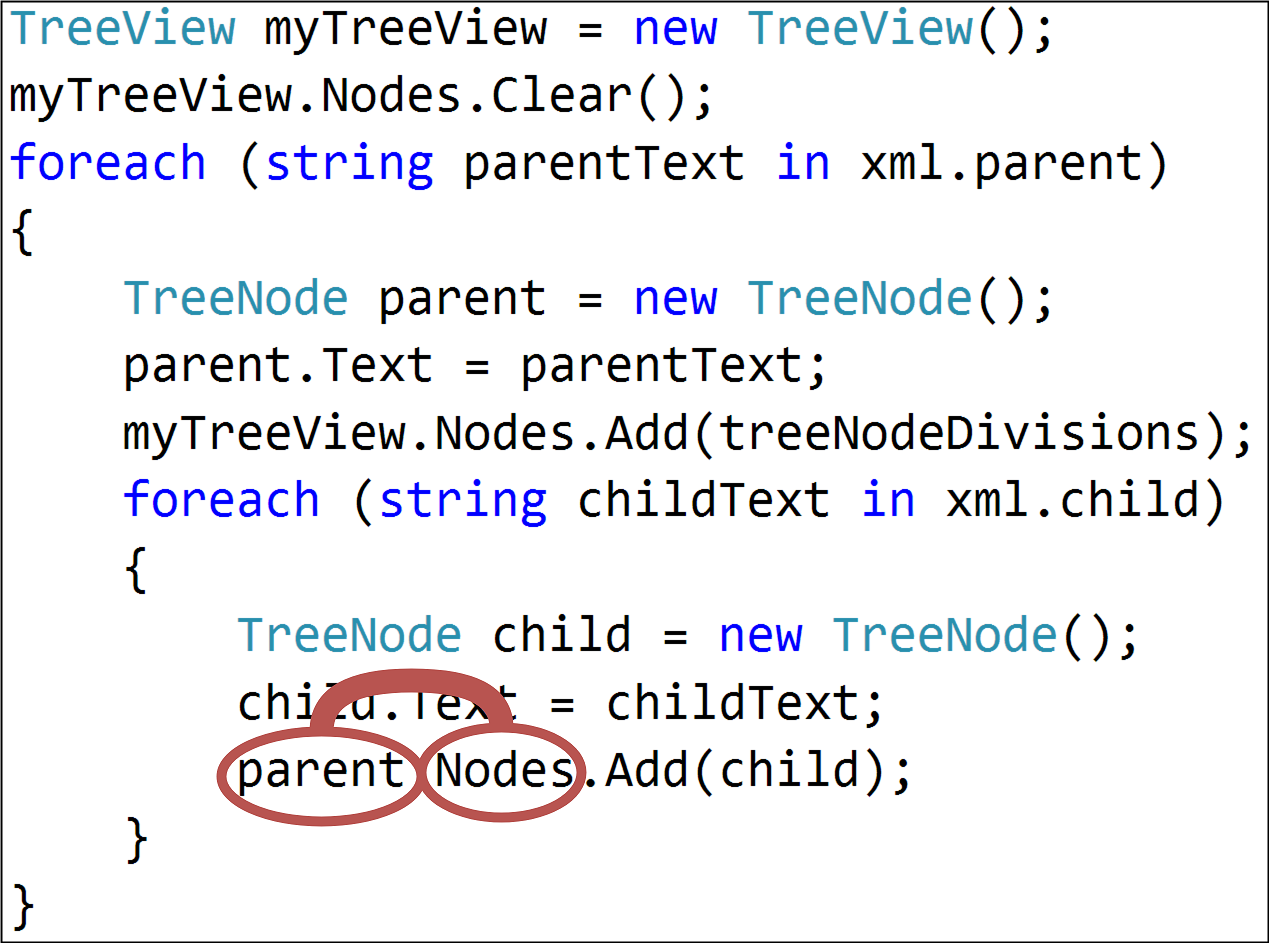}
\end{subfigure}

\begin{subfigure}[t]{\treenodecaptionwidth}
{\raisebox{\raisecaption}{
{\colorbox{white}{\framebox{\large \textbf{c}}}}
}}
\end{subfigure}

\begin{subfigure}[t]{\treenodefigurewidth}

\includegraphics[height=\treenodeheight,keepaspectratio]{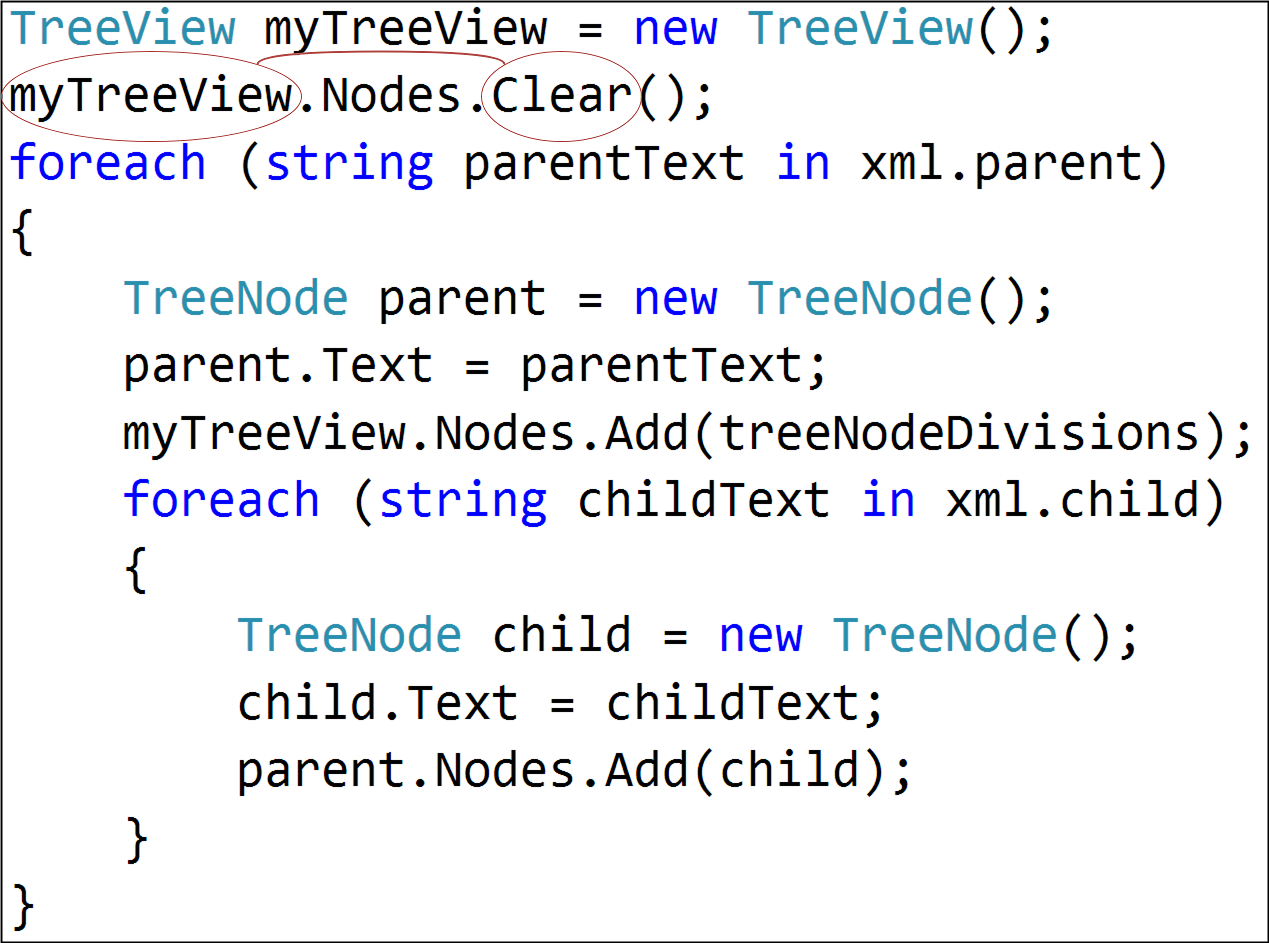}
\end{subfigure}

\begin{subfigure}[t]{\treenodecaptionwidth}
{\raisebox{\raisecaption}{
{\colorbox{white}{\framebox{\large \textbf{\#}}}}
}}
\end{subfigure}
\\
\begin{subfigure}[t]{\textwidth}
\centering
\framebox{\Large \textbf{add a child node to a treeview in c \#}}
\end{subfigure}
\end{tabular}
\end{minipage}
\caption{Example of code captioning for a C\# code snippet from our test set. The text boxes at the bottom of each figure are the predictions produced by our model at each decoding step. The highlighted paths in each figure are the top-attended paths in each decoding step, and their widths are proportional to their attention weight (because of space limitations we included only the top-attended path for each decoding step, but hundreds of paths are attended at each step).}
\label{fig:treenode}
\end{figure*} 
\usemintedstyle{vs}
\begin{figure}[t]
\noindent\rule{\textwidth}{1pt}
\centering

\vspace{2mm}
\begin{subfigure}[t!]{0.8\textwidth}
\begin{minted}[fontsize=\footnotesize, frame=single,framesep=2pt]{csharp}
TreeView myTreeView = new TreeView();
myTreeView.Nodes.Clear();
foreach (string parentText in xml.parent)
{
  TreeNode parent = new TreeNode();
  parent.Text = parentText;
  myTreeView.Nodes.Add(treeNodeDivisions);

  foreach (string childText in xml.child)
  {
    TreeNode child = new TreeNode();
    child.Text = childText;
    parent.Nodes.Add(child);
  }
}
\end{minted}
\end{subfigure}
\\
\vspace{1mm}
\begin{subfigure}[]{0.9\textwidth}
\centering
\begin{tabular}{|l|l|}
\multicolumn{1}{l}{\bf Model}  &\multicolumn{1}{l}{\bf Prediction}
\scriptsize

\\ \hline
MOSES$^{\dag}$ \cite{koehn2007moses}  & \makecell[ll]{
How can TreeView TreeView a TreeView \\
nodes from XML parentText string to a \\
treeview node from a TreeView parentText \\
of a tree treeNodeDivisions from to \\
child childText XML node of MDI child \\
childText created in a tree nodes in }\\
\hline
IR$^{\dag}$                           & How to set the name of a tabPage progragmatically \\
\hline
SUM-NN$^{\dag}$ \cite{rush2015}       & how to get data from xml file in c\# \\
\hline
2-layer BiLSTM         & how to add child nodes to treeview \\
\hline
Transformer \cite{vaswani2017attention}       & how to add child node in treeview in c \# \\
\hline
TreeLSTM \cite{tai2015improved} & how to get the value of a node in xml \\
\hline
CodeNN$^{\dag}$ \cite{codenn16}             & How to get all child nodes in TreeView ? \\
\Xhline{4\arrayrulewidth}
code2seq (this work)             & add a child node to a treeview in c \# \\
\Xhline{4\arrayrulewidth}

\end{tabular}
\end{subfigure} \\
\vspace{2mm}
\noindent\rule{\textwidth}{1pt}
\begin{subfigure}[t!]{0.8\textwidth}
\vspace{2mm}
\begin{minted}[fontsize=\footnotesize, frame=single,framesep=2pt]{csharp}
var excel = new ExcelQueryFactory("excelFileName");
var firstRow = excel.Worksheet().First();
var companyName = firstRow["CompanyName"];
\end{minted}
\end{subfigure}
\vspace{1mm}
\begin{subfigure}[]{0.9\textwidth}
\centering
\begin{tabular}{|l|l|}
\multicolumn{1}{l}{\bf Model}  &\multicolumn{1}{l}{\bf Prediction}
\\ \hline
MOSES$^{\dag}$ \cite{koehn2007moses}  & \makecell[ll]{
How into string based on an firstRow \\
a companyName firstRow ? How to  } \\
\hline
IR$^{\dag}$                           & Facebook C\# SDK Get Current User \\
\hline
SUM-NN$^{\dag}$ \cite{rush2015}       & how can i get the value of a string? \\
\hline
2-layer BiLSTM          & how to get the value of a cell in excel using c \# \\
\hline
Transformer \cite{vaswani2017attention}      & getting the first row in excel \\
\hline
TreeLSTM \cite{tai2015improved} & how to get the value of a cell in excel using c \# \\
\hline
CodeNN$^{\dag}$ \cite{codenn16}             & how do I get the value of an xml file in c \# ? \\
\Xhline{4\arrayrulewidth}
code2seq (this work)             & get the value of a column in excel using c \# \\
\Xhline{4\arrayrulewidth}

\end{tabular}
\end{subfigure} \\
\vspace{2mm}
\noindent\rule{\textwidth}{1pt}
\end{figure}
\begin{figure}[t]
\noindent\rule{\textwidth}{1pt}
\centering
\ContinuedFloat
\begin{subfigure}[t!]{0.85\textwidth}
\vspace{2mm}
\begin{minted}[fontsize=\footnotesize, frame=single,framesep=2pt]{csharp}
static void Main(string[] args)
{
  // Create an instance of Bytescout.PDFRenderer.
  // RasterRenderer object and register it.
  RasterRenderer renderer = new RasterRenderer();
  renderer.RegistrationName = "demo";
  renderer.RegistrationKey = "demo";
  // Load PDF document.
  renderer.LoadDocumentFromFile("multipage.pdf");
  for (int i = 0; i < renderer.GetPageCount(); i++)
  {
    // Render first page of the document to BMP image file.
    renderer.RenderPageToFile(i, RasterOutputFormat.BMP,
      "image" + i + ".bmp");
  }

  // Open the first output file in default image viewer.
  System.Diagnostics.Process.Start("image0.bmp");
}
\end{minted}
\end{subfigure}
\\
\vspace{1mm}
\begin{subfigure}[]{0.9\textwidth}
\centering
\begin{tabular}{|l|l|}
\multicolumn{1}{l}{\bf Model}  &\multicolumn{1}{l}{\bf Prediction}
\scriptsize
\\ \hline
MOSES$^{\dag}$ \cite{koehn2007moses}  & \makecell[ll]{
How to add RasterRenderer renderer \\
RasterRenderer renderer in a string \\
in RegistrationName renderer Registration \\
Key renderer LoadDocumentFromFile in C \# \\
How to a renderer Is a RenderPageToFile \\
renderer in a string to BMP
RasterOutputFormat \\
each in C \# } \\
\hline
IR$^{\dag}$                           & \makecell[ll]{Select TOP 5 * from SomeTable, \\ using Dataview.RowFilter? } \\
\hline
SUM-NN$^{\dag}$ \cite{rush2015}       & how do i create a text file in c\# \\
\hline
2-layer BiLSTM         & how do i create a pdf file in c \# ? \\
\hline
Transformer \cite{vaswani2017attention}      & how to merge two pdf files ? \\
\hline
TreeLSTM \cite{tai2015improved} & how to make a process in c \#\\
\hline
CodeNN$^{\dag}$ \cite{codenn16}             & How to get the value of an array in C \# ? \\
\Xhline{4\arrayrulewidth}
code2seq (this work)             & get the image from a pdf file in c \# \\
\Xhline{4\arrayrulewidth}

\end{tabular}
\end{subfigure} \\
\vspace{2mm}
\noindent\rule{\textwidth}{1pt}

\begin{subfigure}[t!]{0.8\textwidth}
\vspace{2mm}
\begin{minted}[fontsize=\footnotesize, frame=single,framesep=2pt]{csharp}
void Main() {
  string text = File.ReadAllText(@"T:\File1.txt");
  int num = 0;
  text = (Regex.Replace(text, "map", delegate(Match m) {
    return "map" + num++;
  }));
  File.WriteAllText(@"T:\File1.txt", text);
}
\end{minted}
\end{subfigure}
\vspace{1mm}
\begin{subfigure}[]{0.9\textwidth}
\centering
\begin{tabular}{|l|l|}
\multicolumn{1}{l}{\bf Model}  &\multicolumn{1}{l}{\bf Prediction}
\scriptsize
\\ \hline
MOSES$^{\dag}$ \cite{koehn2007moses}  & \makecell[ll]{
How to File then How to HTML ? C \# How to \\
Write to
} \\
\hline
IR$^{\dag}$                           & C\# remove extra carriage returns from Stream \\
\hline
SUM-NN$^{\dag}$ \cite{rush2015}       & how do i create a text file in c\# \\
\hline
2-layer BiLSTM          & how to read a text file from a text file \\
\hline
Transformer \cite{vaswani2017attention}      & how to write a . txt file in c \# \\
\hline
TreeLSTM \cite{tai2015improved} & how to get the text from a text file in c \# \\
\hline
CodeNN$^{\dag}$ \cite{codenn16}             & how to read a text file in c \# ? \\
\Xhline{4\arrayrulewidth}
code2seq (this work)             & replace a string in a text file \\
\Xhline{4\arrayrulewidth}

\end{tabular}
\end{subfigure} \\

\noindent\rule{\textwidth}{1pt}
\caption{C\# examples from our test set for the code captioning task, along with the prediction of our model and each of the baselines.}
\label{appendix_csharp_figure}

\end{figure} 

\begin{figure}[t]
\centering

\noindent\rule{\textwidth}{1pt}
\begin{subfigure}[t!]{0.98\textwidth}
\vspace{2mm}
\begin{minted}[fontsize=\footnotesize, frame=single,framesep=2pt]{Java}
void ______(Counter childCounter, String request, String requestId,
             long duration, boolean systemError, int responseSize) {
  // si je suis le counter fils du counter du contexte parent
  // comme sql pour http alors on ajoute la requête fille
  if (parentContext != null && parentCounter.getName()
    .equals(parentContext.getParentCounter().getChildCounterName())) {
      childHits++;
      childDurationsSum += (int) duration;
    }

  // pour drill-down on conserve pour chaque requête mère, les requêtes
  //  filles appelées et le nombre d'exécutions pour chacune
  if (parentContext == null) {
    addChildRequestForDrillDown(requestId);
  } else {
    parentContext.addChildRequestForDrillDown(requestId);
  }
}
\end{minted}
\end{subfigure}
\\
\vspace{1mm}
\begin{subfigure}[]{0.9\textwidth}
\centering
\begin{tabular}{|l|l|}
\multicolumn{1}{l}{\bf Model}  &\multicolumn{1}{l}{\bf Prediction}
\scriptsize

\\ \hline
ConvAttention \cite{conv16}  & add \\
\hline
Paths+CRFs \cite{pigeon}                           & call \\
\hline
code2vec \cite{alon2019code2vec}       & log response \\
\hline
2-layer BiLSTM (no token splitting)          & handle request\\
\hline
2-layer BiLSTM         & report child request \\
\hline
Transformer       & add child \\
\hline
TreeLSTM \cite{tai2015improved} & add child \\
\Xhline{4\arrayrulewidth}
Gold:                            & add child request \\
\Xhline{4\arrayrulewidth}
code2seq (this work)             &  add child request \\
\Xhline{4\arrayrulewidth}

\end{tabular}
\end{subfigure} \\
\vspace{2mm}
\noindent\rule{\textwidth}{1pt}

\begin{subfigure}[t!]{0.8\textwidth}
\vspace{2mm}
\begin{minted}[fontsize=\footnotesize, frame=single,framesep=2pt]{Java}
public static int ______(int value) {
  return value <= 0 ? 1 :
    value >= 0x40000000 ? 0x40000000 :
      1 << (32 - Integer.numberOfLeadingZeros(value - 1));
}
\end{minted}
\end{subfigure}
\vspace{1mm}
\begin{subfigure}[]{0.9\textwidth}
\centering
\begin{tabular}{|l|l|}
\multicolumn{1}{l}{\bf Model}  &\multicolumn{1}{l}{\bf Prediction}
\\ \hline
ConvAttention \cite{conv16}  & get \\
\hline
Paths+CRFs \cite{pigeon}                           & test bit inolz \\
\hline
code2vec \cite{alon2019code2vec}       & multiply \\
\hline
2-layer BiLSTM (no token splitting)          & next power of two \\
\hline
2-layer BiLSTM         & \{ \qquad\emph{(replaced UNK)} \\
\hline
Transformer       & get bit length \\
\hline
TreeLSTM \cite{tai2015improved} & get \\
\Xhline{4\arrayrulewidth}
Gold:                            & find next positive power of two \\
\Xhline{4\arrayrulewidth}
code2seq (this work)             &  get power of two \\
\Xhline{4\arrayrulewidth}

\end{tabular}
\end{subfigure} \\
\vspace{2mm}
\noindent\rule{\textwidth}{1pt}
\end{figure}


\begin{figure}[t]
\centering

\noindent\rule{\textwidth}{1pt}

\begin{subfigure}[t!]{0.95\textwidth}
\vspace{2mm}
\begin{minted}[fontsize=\footnotesize, frame=single,framesep=2pt]{Java}
BigInteger ______(int bitlength, BigInteger e, BigInteger sqrdBound)
{
  for (int i = 0; i != 5 * bitlength; i++)
  {
    BigInteger p = new BigInteger(bitlength, 1, param.getRandom());
    if (p.mod(e).equals(ONE))
    {
      continue;
    }
    if (p.multiply(p).compareTo(sqrdBound) < 0)
    {
      continue;
    }
    if (!isProbablePrime(p))
    {
      continue;
    }
    if (!e.gcd(p.subtract(ONE)).equals(ONE))
    {
      continue;
    }
    return p;
  }
  throw new IllegalStateException("unable to generate prime number..
    ...for RSA key");
}
\end{minted}
\end{subfigure}
\\
\vspace{1mm}
\begin{subfigure}[]{0.9\textwidth}
\centering
\begin{tabular}{|l|l|}
\multicolumn{1}{l}{\bf Model}  &\multicolumn{1}{l}{\bf Prediction}
\scriptsize

\\ \hline
ConvAttention \cite{conv16}  &  test\\
\hline
Paths+CRFs \cite{pigeon}                           & i \\
\hline
code2vec \cite{alon2019code2vec}       & to big integer \\
\hline
2-layer BiLSTM (no token splitting)          & generate prime \\
\hline
2-layer BiLSTM          & generate prime number \\
\hline
Transformer       & generate\\
\hline
TreeLSTM \cite{tai2015improved} & probable prime \\
\Xhline{4\arrayrulewidth}
Gold:                            & choose random prime \\
\Xhline{4\arrayrulewidth}
code2seq (this work)             & generate prime number  \\
\Xhline{4\arrayrulewidth}

\end{tabular}
\end{subfigure} \\
\vspace{2mm}
\noindent\rule{\textwidth}{1pt}

\begin{subfigure}[t!]{0.75\textwidth}
\vspace{2mm}
\begin{minted}[fontsize=\footnotesize, frame=single,framesep=2pt]{Java}
public boolean ______(Set<String> set, String value) {
  for (String entry : set) {
    if (entry.equalsIgnoreCase(value)) {
      return true;
    }
  }
  return false;
}
\end{minted}
\end{subfigure}
\vspace{1mm}
\begin{subfigure}[]{0.9\textwidth}
\centering
\begin{tabular}{|l|l|}
\multicolumn{1}{l}{\bf Model}  &\multicolumn{1}{l}{\bf Prediction}
\\ \hline
ConvAttention \cite{conv16}  & is \\
\hline
Paths+CRFs \cite{pigeon}                           & equals \\
\hline
code2vec \cite{alon2019code2vec}       & contains ignore case \\
\hline
2-layer BiLSTM (no token splitting)          & contains ignore case \\
\hline
2-layer BiLSTM         & contains \\
\hline
Transformer      &  contains \\
\hline
TreeLSTM \cite{tai2015improved} & contains ignore case \\
\Xhline{4\arrayrulewidth}
Gold:                            & contains ignore case \\
\Xhline{4\arrayrulewidth}
code2seq (this work)             & contains ignore case \\
\Xhline{4\arrayrulewidth}

\end{tabular}
\end{subfigure} \\
\vspace{2mm}
\noindent\rule{\textwidth}{1pt}
\caption{Java examples from our test set for the code summarization task, along with the prediction of our model and each of the baselines.}
\label{appendix_java_figure}
\end{figure} 

\section{Code Captioning results}\label{csharp_appendix_section}
\definecolor{treelstmcolor}{HTML}{0F9D58}
\definecolor{bilstmfullcolor}{HTML}{000080}
\definecolor{bilstmsplitcolor}{HTML}{FF6d00}
\definecolor{transformercolor}{HTML}{46BDC6}
\definecolor{c2scolor}{HTML}{AB30C4}

\definecolor{mosescolor}{HTML}{4285F4}
\definecolor{ircolor}{HTML}{DB4437}
\definecolor{sumnncolor}{HTML}{F4B400}
\definecolor{codenncolor}{HTML}{000080}

\begin{figure}
\centering
\begin{tikzpicture}
\begin{axis}[
    ybar,
    bar width=16pt,
    width=0.5\textwidth,
    height=7cm,
    enlarge x limits=0.10,
    ylabel={BLEU},
    symbolic x coords={StackOverflow dataset},
    xtick=data,
    x tick label style  = {font=\normalsize},
    nodes near coords style={
        color=black,
        font=\tiny
    },%
    nodes near coords,
    font=\scriptsize,
    legend style={font=\normalsize,at={(1.01,0.5)},
    anchor=west, row sep=0.5pt},
    ytick={0.0,5.0,...,25.0},
    ymin=0,
    grid = major,
    grid style={line width=0.1pt, draw=gray!10},
    major grid style={line width=.2pt,draw=gray!50},
    ylabel style={rotate=-90,font=\normalsize},
    ]
\addplot[mosescolor,fill=mosescolor] coordinates {(StackOverflow dataset,11.57)};
\addplot[ircolor,fill=ircolor] coordinates {(StackOverflow dataset,13.66)};
\addplot[sumnncolor,fill=sumnncolor] coordinates {(StackOverflow dataset,19.31)};
\addplot[bilstmsplitcolor,fill=bilstmsplitcolor] coordinates {(StackOverflow dataset,19.78)};
\addplot[transformercolor,fill=transformercolor] coordinates {(StackOverflow dataset,19.68)};
\addplot[treelstmcolor,fill=treelstmcolor] coordinates {(StackOverflow dataset,20.11)};
\addplot[codenncolor,fill=codenncolor] coordinates {(StackOverflow dataset,20.53)};
\addplot[c2scolor,fill=c2scolor] coordinates {(StackOverflow dataset,23.04)};
\legend{
    MOSES$^{\dag}$ \cite{koehn2007moses},
    IR$^{\dag}$,
    SUM-NN$^{\dag}$ \cite{rush2015},
    2-layer BiLSTM,
    Transformer \cite{vaswani2017attention},
    TreeLSTM \cite{tai2015improved},
    CodeNN$^{\dag}$ \cite{codenn16},
    \textbf{code2seq} (this work)
}
\end{axis}
\end{tikzpicture}
\caption{Visualization of the BLEU score of our model compared to the baselines, for the code captioning task. The values are the same as in \Cref{csharp-results}. Our model achieves significantly higher results than the baselines.}
\label{csharp_results_barchart_figure}
\end{figure}
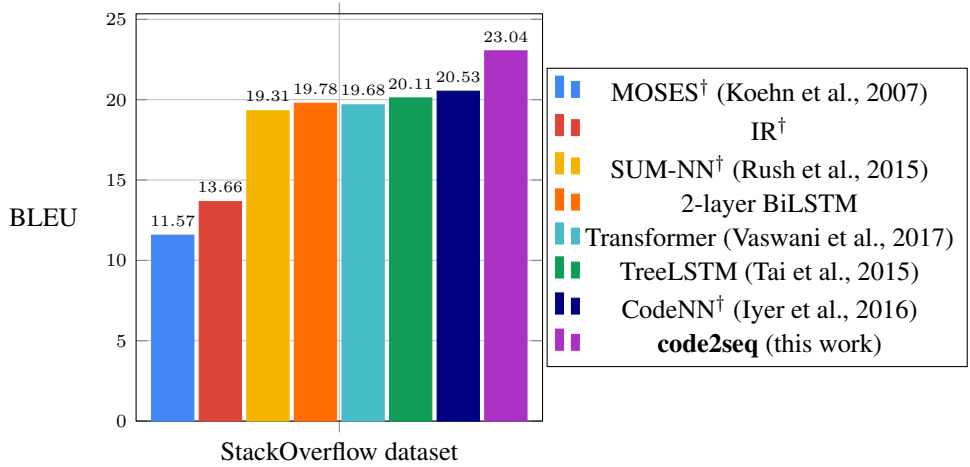 
\Cref{csharp_results_barchart_figure} shows a bar chart of the BLEU score of our model and the baselines, in the code captioning task (predicting natural language descriptions for C\# code snippets). The numbers are the same as in \Cref{csharp-results}.

\section{Code Summarization results}\label{java_appendix_section}
\definecolor{convcolor}{HTML}{4285F4}
\definecolor{pigeoncolor}{HTML}{DB4437}
\definecolor{c2vcolor}{HTML}{F4B400}
\definecolor{treelstmcolor}{HTML}{0F9D58}
\definecolor{bilstmfullcolor}{HTML}{000080}
\definecolor{bilstmsplitcolor}{HTML}{FF6d00}
\definecolor{transformercolor}{HTML}{46BDC6}
\definecolor{c2scolor}{HTML}{AB30C4}

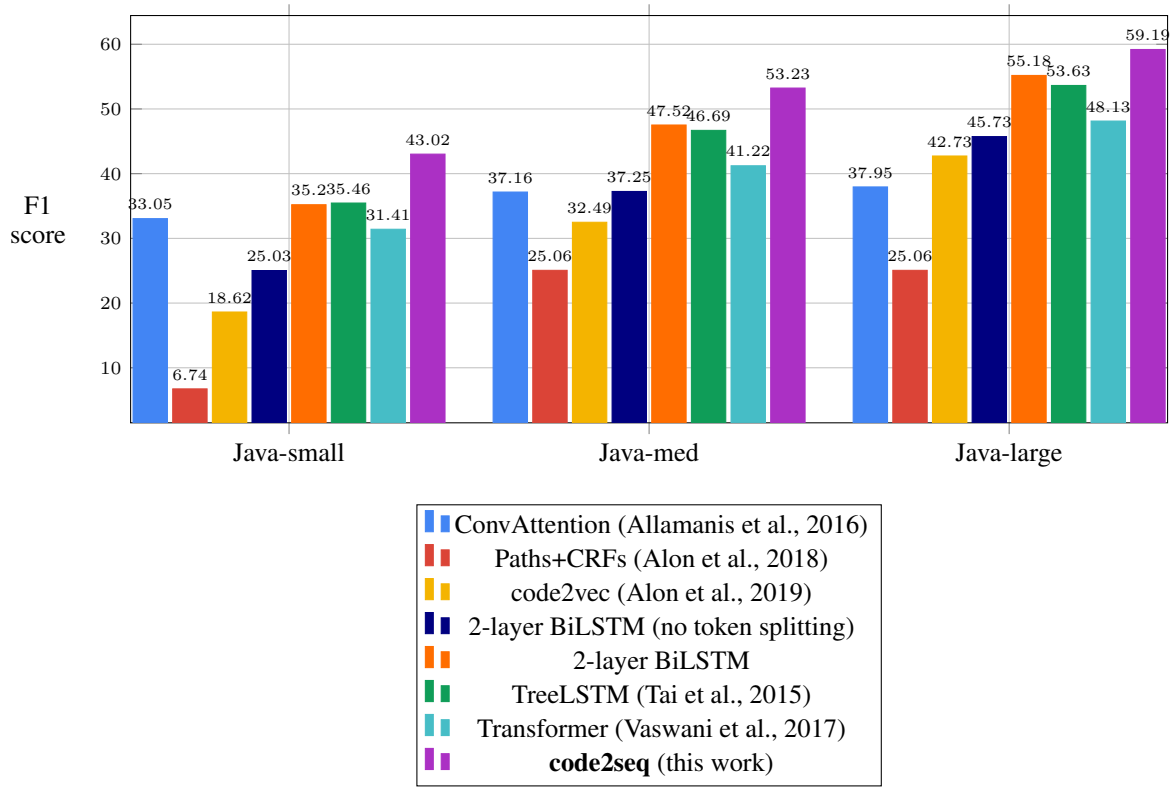
\begin{figure}[t]
\hspace{-10mm}
\begin{tikzpicture}
\begin{axis}[
    ybar,
    bar width=13pt,
    width=1.1\textwidth,
    height=7cm,
    enlarge x limits=0.22,
    ylabel style={align=center},
    ylabel=F1 \\score,
    x tick label style  = {font=\normalsize},
    symbolic x coords={Java-small, Java-med, Java-large},
    xtick=data,
    nodes near coords style={
        color=black,
        font=\tiny
    },%
    nodes near coords,
    font=\scriptsize,
    legend style={font=\normalsize,at={(0.5,-0.2)},
    anchor=north, row sep=0.5pt},
    ytick={0.0,10.0,...,60.0},
    grid = major,
    grid style={line width=0.1pt, draw=gray!10},
    major grid style={line width=.2pt,draw=gray!50},
    ylabel style={rotate=-90, font=\normalsize},
    ]
\addplot[convcolor,fill=convcolor] coordinates {(Java-small,33.05) (Java-med,37.16) (Java-large,37.95)};
\addplot[pigeoncolor,fill=pigeoncolor] coordinates {(Java-small,6.74) (Java-med,25.06) (Java-large,25.06)};
\addplot[c2vcolor,fill=c2vcolor] coordinates {(Java-small,18.62) (Java-med,32.49) (Java-large,42.73)};
\addplot[bilstmfullcolor,fill=bilstmfullcolor] coordinates {(Java-small,25.03) (Java-med,37.25) (Java-large,45.73)};
\addplot[bilstmsplitcolor,fill=bilstmsplitcolor] coordinates {(Java-small,35.20) (Java-med,47.52) (Java-large,55.18)};
\addplot[treelstmcolor,fill=treelstmcolor] coordinates {(Java-small,35.46) (Java-med,46.69) (Java-large,53.63)};
\addplot[transformercolor,fill=transformercolor] coordinates {(Java-small,31.41) (Java-med,41.22) (Java-large,48.13)};
\addplot[c2scolor,fill=c2scolor] coordinates {(Java-small,43.02) (Java-med,53.23) (Java-large,59.19)};
\legend{ConvAttention \cite{conv16}, Paths+CRFs \cite{pigeon}, code2vec \cite{alon2019code2vec}, 2-layer BiLSTM (no token splitting), 2-layer BiLSTM, TreeLSTM \cite{tai2015improved}, Transformer \cite{vaswani2017attention}, \textbf{code2seq} (this work)}
\end{axis}
\end{tikzpicture}
\caption{Visualization of the F1 score of our model compared to the baselines, for the code summarization task, across datasets. The values are the F1 columns from \Cref{methods-results}. Our model achieves significantly higher results than the baselines.}
\label{java_results_barchart_figure}
\end{figure}
\Cref{java_results_barchart_figure} shows a bar chart of the F1 score of our model and the baselines, in the code summarization task (predicting method names in Java). The numbers are the F1 columns from \Cref{methods-results}.

\section{Ablation study results}\label{ablation_appendix_section}
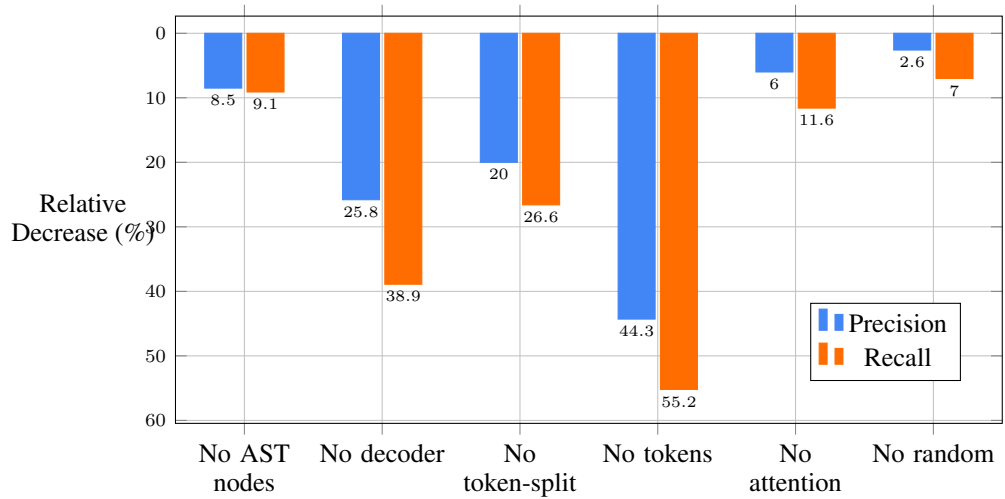
\begin{figure}
\begin{tikzpicture}
\begin{axis}[
    ybar,
    bar width=14pt,
    width=0.9\textwidth,
    height=7cm,
    enlarge x limits=0.10,
    ylabel style={align=center},
    ylabel=Relative\\Decrease (\%),
    x tick label style  = {text width=1.7cm,align=center,font=\normalsize},
    symbolic x coords={No AST nodes,No decoder,No token-split,No tokens,No attention,No random},
    xtick=data,
    nodes near coords style={
        color=black,
        font=\tiny,
        yshift=-10pt
    },%
    nodes near coords,
    font=\scriptsize,
    legend style={font=\normalsize,at={(0.95,0.2)},
    anchor=east, row sep=0.5pt},
    ytick={0.0,10.0,...,60.0},
    grid = major,
    grid style={line width=0.1pt, draw=gray!10},
    major grid style={line width=.2pt,draw=gray!50},
    ylabel style={rotate=-90, font=\normalsize},
    y dir=reverse,
    ]
\addplot[convcolor,fill=convcolor] coordinates {(No AST nodes,8.5) (No decoder,25.8) (No token-split,20.0) (No tokens,44.3) (No attention,6.0) (No random,2.6)};
\addplot[bilstmsplitcolor,fill=bilstmsplitcolor] coordinates {(No AST nodes,9.1) (No decoder,38.9) (No token-split,26.6) (No tokens,55.2) (No attention,11.6) (No random,7.0)};
\legend{Precision, Recall}
\end{axis}
\end{tikzpicture}
\caption{The relative decrease in precision and recall in each of the ablations, compared to the full model.}
\label{ablation_barchart_figure}
\end{figure}
\Cref{ablation_barchart_figure} shows a bar chart of the relative decrease in precision and recall for each of the ablations described in \Cref{sec:analysis} and presented in \Cref{ablation-results}.

\end{document}